\documentclass[journal]{IEEEtran}

\ifCLASSINFOpdf

\else

\fi
\usepackage{cite}
\usepackage[square,sort&compress,numbers]{natbib}
\usepackage{graphicx}
\usepackage{makecell}
\graphicspath{{images/}}

\usepackage{enumitem}
\usepackage{amsmath}
\usepackage{subfigure}
\usepackage{float}
\usepackage{booktabs}
\usepackage{url}        

\usepackage{jabbrv}
\usepackage{textcomp}
\usepackage{multirow}
\usepackage{amsfonts}
\usepackage{mathrsfs}
\usepackage{bbding}
\usepackage{algorithm}
\usepackage{algorithmicx}
\usepackage{algpseudocode}

\usepackage{color,soul}
\soulregister\cite7
\soulregister\ref7
\usepackage{soul}
\soulregister\cite7
\soulregister\ref7 

\usepackage{bm}
\usepackage{hyperref}
\hypersetup{hidelinks, colorlinks=true, allcolors=black, pdfstartview=Fit,	breaklinks=true}

\begin{document}
\title{
SeG-SR: Integrating Semantic Knowledge into Remote Sensing Image Super-Resolution via Vision-Language Model}

\author{Bowen Chen, Keyan Chen, Mohan Yang, Zhengxia Zou,~\IEEEmembership{Senior Member,~IEEE}, and Zhenwei Shi$^\star$,~\IEEEmembership{Senior Member,~IEEE} 

\thanks{
	The work was supported by the National Natural Science Foundation of China under Grants 62125102, U24B20177 and 623B2013, the Beijing Natural Science Foundation under Grant JL23005, and the Fundamental Research Funds for the Central Universities. $^\star$Corresponding author: Zhenwei Shi (e-mail: shizhenwei@buaa.edu.cn).}

\thanks{
	Bowen Chen, Keyan Chen, Mohan Yang, Zhengxia Zou and Zhenwei Shi are with the Department of Aerospace Intelligent Science and Technology, School of Astronautics, Beihang University, Beijing 100191, China; and Key Laboratory of Spacecraft Design Optimization and Dynamic Simulation Technologies, Ministry of Education, Beihang University, Beijing 100191, China.}
}

%

\maketitle




\begin{abstract}
High-resolution (HR) remote sensing imagery plays a vital role in a wide range of applications, including urban planning and environmental monitoring. However, due to limitations in sensors and data transmission links, the images acquired in practice often suffer from resolution degradation. Remote Sensing Image Super-Resolution (RSISR) aims to reconstruct HR images from low-resolution (LR) inputs, providing a cost-effective and efficient alternative to direct HR image acquisition. 
Existing RSISR methods primarily focus on low-level characteristics in pixel space, while neglecting the high-level understanding of remote sensing scenes. This may lead to semantically inconsistent artifacts in the reconstructed results. Motivated by this observation, our work aims to explore the role of high-level semantic knowledge in improving RSISR performance. We propose a Semantic-Guided Super-Resolution framework, SeG-SR, which leverages Vision-Language Models (VLMs) to extract semantic knowledge from input images and uses it to guide the super resolution (SR) process. Specifically, we first design a Semantic Feature Extraction Module (SFEM) that utilizes a pretrained VLM to extract semantic knowledge from remote sensing images. Next, we propose a Semantic Localization Module (SLM), which derives a series of semantic guidance from the extracted semantic knowledge. Finally, we develop a Learnable Modulation Module (LMM) that uses semantic guidance to modulate the features extracted by the SR network, effectively incorporating high-level scene understanding into the SR pipeline. 
We validate the effectiveness and generalizability of SeG-SR through extensive experiments: SeG-SR achieves state-of-the-art performance on three datasets, and consistently improves performance across various SR architectures. Notably, for the $\times$4 SR task on the UCMerced dataset, it attained a PSNR of 29.3042 dB and an SSIM of 0.7961.
Codes can be found at \href{https://github.com/Mr-Bamboo/SeG-SR}{https://github.com/Mr-Bamboo/SeG-SR}. 

\end{abstract}


\IEEEpeerreviewmaketitle

\section{Introduction}

\IEEEPARstart{H}{igh}-resolution (HR) remote sensing images play an significant role in numerous applications, such as land cover mapping \cite{chen2024rsprompter, yuan2020deep, tong2020land}, target detection \cite{chen2023target, wang2022remote, chen2025dynamicvis}, and change detection \cite{liu2024rscama, zhang2024bifa, zhang2025cdmamba, liu2022cc}. However, due to constraints related to sensors and transmission links, remote sensing images obtained in practice are typically of low-resolution (LR), lacking sufficient spatial detail and consequently inadequate for fine-grained remote sensing interpretation \cite{chen2024spectral, chen2024leveraging}. Therefore, enhancing the spatial resolution of remote sensing images is critically important for subsequent remote sensing tasks. 

Super-resolution (SR) provides a computational approach to increasing the spatial resolution of remote sensing images \cite{xiao2023ediffsr,lei2021transformer}. Compared to hardware upgrades, SR algorithms represent a more economical and flexible solution. Early SR approaches primarily relied on traditional machine learning techniques, such as Markov random fields \cite{freeman2002example}, neighborhood embedding \cite{chang2004super}, sparse coding \cite{yang2010image, zeyde2010single, hou2017adaptive}, and random forests \cite{schulter2015fast}. These methods are often characterized by complex optimization processes and generally fail to achieve satisfactory reconstruction quality \cite{yang2019deep}. \color{black} In contrast, deep learning-based SR methods leverage neural networks to automatically learn complex mappings between LR and HR image pairs \cite{lei2017super, dong2015image}, effectively reconstructing detailed information and thus gradually becoming dominant in remote sensing image SR (RSISR) research \cite{xiao2024frequency, sui2023dtrn}.

However, remote sensing images present complex scene characteristics, specifically:

(1) Ground objects exhibit significant diversity \cite{liu2023diverse}. Remote sensing images typically encompass diverse types of ground objects, and even within the same object type, substantial differences in shape and texture may exist.

(2) Complex coupling relationships among ground objects. Unlike natural images, which often feature distinct foreground subjects, remote sensing images usually lack clear distinctions between foreground and background.

Consequently, RSISR requires specialized designs to adequately address intricate details and relational complexities of ground objects. Previous methods have primarily relied on prior assumptions about the low-level characteristics (such as texture) of remote sensing images based on the empirical observations. For instance, Lei et al. \cite{lei2021hybrid} observed the repetitive nature of ground objects in remote sensing scenes and addressed complex textures by modeling self-similarity. Wu et al. \cite{wu2023lightweight} introduced a saliency detection module, treating foreground regions as highly salient areas and applying more complex reconstruction strategies to them. Chen et al. \cite{chen2025heterogeneous} employed a Mixture-of-Experts (MoE) \cite{riquelme2021scaling} model during the upsampling stage, where experts with different receptive fields were used to reconstruct diverse types of details. However, since these methods only focus on low-level features during the SR process and lack high-level understanding of remote sensing scenes, the reconstructed results often suffer from semantically inconsistent artifacts. Although some approaches have attempted to incorporate scene-level understanding through multi-task learning frameworks that jointly train semantic segmentation and super-resolution \cite{wang2020dual, zhang2022collaborative}, such methods typically rely on manual semantic annotations and may suffer from conflicting optimization objectives across tasks \cite{sener2018multi}.\color{black}

\begin{figure}[t]
    \centering    
    \includegraphics[width=0.5\textwidth]{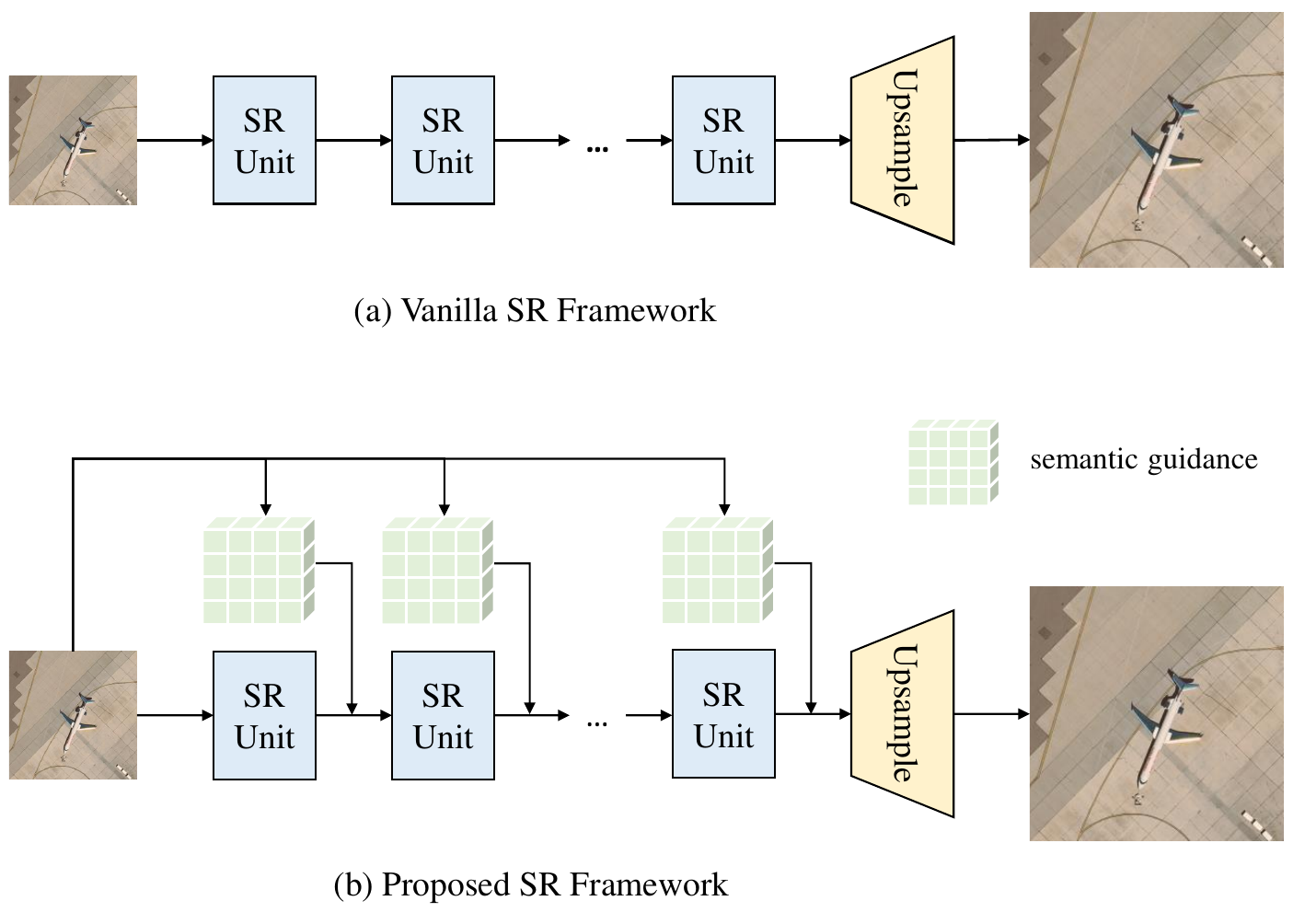}	
    \caption{The previous vanilla SR framework is illustrated in (a), while our proposed SR framework is shown in (b). Our framework introduces semantic guidance information for each SR unit, thereby guiding the SR process.}
	\label{fig:new framework}
\end{figure}

To address this issue, we propose a \textbf{Se}mantic-\textbf{G}uided \textbf{S}uper-\textbf{R}esolution framework, termed SeG-SR, as illustrated in Fig. \ref{fig:new framework}. Unlike the conventional RSISR methods, our framework leverages semantic-level knowledge to guide the super-resolution process. Specifically, we introduce a series of semantic guidance information to identify critical semantic regions within each unit of the SR network. These semantic regions subsequently serve as modulation signals, effectively guiding the SR reconstruction process. Notably, SeG-SR is a generic and plug-and-play framework that can be seamlessly integrated into various SR models, endowing them with high-level semantic understanding of remote sensing imagery. 


Recently, Vision-Language Models (VLMs) have demonstrated remarkable capabilities in capturing semantic and contextual information within images \cite{zhang2024vision}. Typically pretrained on large-scale image-text pairs, VLMs significantly surpass purely visual models in understanding image semantics. Inspired by this observation, we adopt the classic VLM, Contrastive Language-Image Pre-Training (CLIP) \cite{radford2021learning}, to extract semantic knowledge from the input LR image and design Semantic Feature Extraction Module (SFEM) for this purpose. Subsequently, to generate highly targeted semantic guidance that informs the entire SR process, we further propose a Semantic Localization Module (SLM). This module employs a set of learnable embeddings to query the semantic features extracted by CLIP and produce a series of guidance maps, each tailored to a specific SR unit. Finally, these semantic maps are passed through a Learnable Modulation Module (LMM), which modulates the output features of each SR unit. This modulation mechanism enables the network to focus on semantically relevant regions.  Experimental results demonstrate that SeG-SR achieves state-of-the-art performance on the UCMerced \cite{yang2010bag} dataset, the AID \cite{xia2017aid} dataset and the SIRI-WHU dataset \cite{zhao2016fisher, zhao2016dirichlet, zhu2016bag}.\color{black} 

The contributions of our work are summarized as follows:

\begin{enumerate}
    \item We propose a semantic-guided SR framework for RSISR task, termed SeG-SR. By leveraging semantic information to guide the SR process, SeG-SR effectively mitigates semantically inconsistent artifacts during detail reconstruction, leading to more accurate results. \color{black}
    
    \item  We introduce VLM into the RSISR task to extract high-level semantic features and fully exploit its potential through three key components: Semantic Feature Extraction Module, Semantic Localization Module, and Learnable Modulation Module.
    
    \item Experimental results on the UCMerced, AID and SIRI-WHU datasets demonstrate that our proposed SeG-SR achieves state-of-the-art performance compared to existing SR methods. \color{black} Furthermore, by reconstructing representative SR models with different architectures using the SeG-SR framework, we validate its generality and adaptability across diverse network designs.
\end{enumerate}

The rest of the paper is organized as follows. In Section \ref{related work}, we introduce SR methods, RSISR methods and VLMs. Section \ref{method} details the SeG-SR method. Section \ref{experiment} provides experimental evaluations on the reconstruction quality. Finally, we draw conclusions in Section \ref{conclusion}.

\section{Related Works}\label{related work}

In this section, we briefly review the super-resolution methods, remote sensing image super-resolution methods, and vision-language models.

\subsection{Super-resolution Method} \label{sr}

Early super-resolution (SR) approaches can be categorized into interpolation-based, reconstruction-based, and learning-based methods \cite{yang2019deep}. Interpolation-based methods \cite{keys2003cubic} enlarge image size through simple computations performed in a single step; however, this approch leads to limited  accuracy. Reconstruction-based methods \cite{sun2008image, yan2015single} typically leverage sophisticated prior knowledge to reconstruct image details, but this often results in substantial computational complexity. Learning-based methods employ traditional machine learning techniques. Freeman et al. \cite{freeman2002example} employed Markov random fields to probabilistically model the relationship between LR and HR patches. Chang et al. \cite{chang2004super} proposed a neighbor embedding method, which reconstructs patch representations by exploiting the local geometric structure in feature space. Yang et al. \cite{yang2010image} further improved SR performance through sparse representation. Schulter et al. \cite{schulter2015fast} introduced random forests into the SR task, thereby simplifying the parameter tuning process. However, these methods often involved complex optimization procedures and generally failed to deliver satisfactory reconstruction quality.\color{black}

Dong et al. \cite{dong2015image} first introduced convolutional neural networks (CNNs) to the SR task, achieving impressive results and setting the stage for deep learning-based approaches to gradually dominate the field \cite{ledig2017photo}. Kim et al. subsequently proposed VDSR \cite{kim2016accurate}, which utilized residual learning to increase network depth, thereby improving reconstruction accuracy. Wang et al. introduced RRDBNet \cite{wang2018esrgan}, leveraging dense connections to further enhance network performance. The emergence of attention mechanisms has infused new vitality into SR research. For instance, RCAN \cite{zhang2018image} incorporated channel attention modules to improve feature learning, while Dai et al. designed Second-order Channel Attention \cite{dai2019second} to further advance feature modeling capabilities.

Recently, with significant breakthroughs of Transformer architectures in computer vision, researchers have begun exploring Transformer-based SR methods. Liang et al. adapted the Swin Transformer architecture to the SR task \cite{liang2021swinir}, achieving notable improvements. Zhou et al. proposed replacing the standard attention mechanism in Transformers with Permuted Self-Attention \cite{zhou2023srformer}, which expanded the receptive field while simultaneously reducing computational overhead. Furthermore, Chen et al. introduced HAT \cite{chen2023activating}, integrating multiple attention mechanisms to effectively exploit the full potential of Transformers for SR.

\subsection{Remote Sensing Image Super-resolution Method} \label{rsisr}

Early super-resolution (SR) methods for remote sensing imagery primarily employed sparse representation techniques \cite{zeyde2010single, hou2017adaptive}. Lei et al. were among the first to introduce CNNs to remote sensing image SR, integrating both local and global representations to model relationships between HR and LR images \cite{lei2017super}. Subsequently, researchers designed increasingly specialized SR methods tailored to the unique characteristics of remote sensing imagery. For instance, Qin et al. \cite{qin2020remote} incorporated gradient maps to guide models focusing on richer edge regions within remote sensing images. Dong et al. \cite{dong2020remote} proposed a second-order learning strategy to effectively capture multi-scale features in remote sensing imagery. Lei et al. \cite{lei2021hybrid} revisited the intrinsic similarity of ground objects in remote sensing images, employing non-local attention mechanisms to extract mixed-scale self-similarity information. Wu et al. \cite{wu2023lightweight} introduced a saliency detection-based framework to classify image patches according to their complexity, applying distinct reconstruction strategies accordingly. Chen et al. \cite{chen2025heterogeneous} developed a MoE-based reconstruction approach, enabling adaptive processing of different ground object details.

In recent years, Transformer-based SR methods have emerged as a mainstream approach for remote sensing imagery due to their strong contextual modeling capabilities. Lei et al. proposed TransENet \cite{lei2021transformer}, utilizing Transformers for multi-scale feature fusion. Hao et al. proposed SPT \cite{hao2024scale}, which integrates a back-projection strategy with Transformers, utilizing pyramid pooling to learn scale-aware low-resolution features and employing back-projection for effective image reconstruction. Xiao et al. presented TTST \cite{xiao2024ttst}, introducing a token selection mechanism that removes redundant features from Transformer-based SR methods, thus improving computational efficiency.

However, existing remote sensing image SR (RSISR) methods typically emphasize low-level features, such as textures, neglecting semantic understanding. This oversight often results in semantic distortions in the reconstructed images. To address this limitation, our work introduces semantic guidance into the SR process by employing vision-language models (VLMs), effectively integrating semantic understanding to enhance reconstruction fidelity.

Recently, some researchers have explored the use of generative models for RSISR, such as generative adversarial networks (GANs)\cite{goodfellow2014generative, wang2021real} and diffusion models \cite{ho2020denoising}. GANs generate images through an adversarial training process between a generator and a discriminator. Lei et al. proposed CDGAN \cite{lei2020coupled}, which enhances the perceptual quality of reconstructed images by designing a coupled discriminator. MSWAGAN \cite{wang2024MSWAGAN} combines Transformer and CNN, introducing multi-scale sliding-window attention to capture textures at multiple scales without increasing parameter count. SRADSGAN \cite{meng2023single} integrates dense sampling and residual learning to improve reconstruction performance at large magnification factors. Diffusion models, on the other hand, progressively generate high-quality images by iteratively denoising random noise through a learned reverse diffusion process. Han et al. \cite{han2023enhancing} introduced diffusion models into RSISR and achieved better perceptual reconstruction quality than GAN-based methods. Xiao et al. \cite{xiao2023ediffsr} designed an efficient diffusion model that improves computational efficiency while enhancing reconstruction quality. Zhu et al. \cite{zhu2025taming} exploited a large pretrained diffusion model as a generative prior, significantly improving reconstruction performance.

However, these generative approaches may hallucinate visually plausible but factually incorrect details \cite{el2022bigprior}. Since our objective is to faithfully restore the details lost due to resolution degradation in remote sensing images, such generative methods are beyond the scope of our study.
\color{black}

\subsection{Vision-Language Models} \label{vlms}

Vision-Language Models (VLMs) are an extension of Large Language Models (LLMs) to visual domains \cite{liu2024change} and aim to learn rich visual-linguistic associations from large-scale image-text pairs. Consequently, these models demonstrate strong generalization and zero-shot inference capabilities across various vision tasks, including image classification, object detection \cite{zhao2022exploiting}, and semantic segmentation \cite{chen2025rsrefseg}. The Contrastive Language-Image Pre-training (CLIP) \cite{radford2021learning} model is one of the most representative and influential VLMs. CLIP comprises a text encoder and an image encoder, trained using a contrastive learning strategy to map textual and visual features into a shared embedding space. Due to its robust representation capabilities, CLIP has achieved remarkable success across diverse tasks such as image classification, semantic segmentation, and object detection.

Since the emergence of Stable Diffusion (SD) \cite{rombach2022high}, CLIP has also been increasingly integrated into image generation tasks \cite{liu2025text2earth, podell2023sdxl}. Specifically, SD employs CLIP’s text encoder to generate textual embeddings closely aligned with image features for input text, subsequently using these embeddings as inputs to guide the text-to-image diffusion process. Inspired by this approach, recent works have begun incorporating VLMs into natural image super-resolution (SR). For example, Yang et al. proposed PASD \cite{yang2024pixel}, initially utilizing high-level vision models to generate image content descriptions, which are then transformed by CLIP's text encoder into embeddings for guiding SD. Similarly, Sun et al. developed CoSeR \cite{sun2024coser}, where an additional reference image is input into CLIP’s image encoder to produce image embeddings matched with textual content, effectively constraining style and detail in the generated images from SD.

However, these methods are primarily designed around SD, a pre-trained generative model inherently prone to fabricating unrealistic details—an outcome unacceptable in remote sensing applications. Moreover, these approaches typically require additional inputs such as reference images or descriptive textual prompts, which are rarely available or practical in the context of remote sensing image SR tasks.

\begin{figure*}
 	\centering
 	\includegraphics[width=\textwidth]{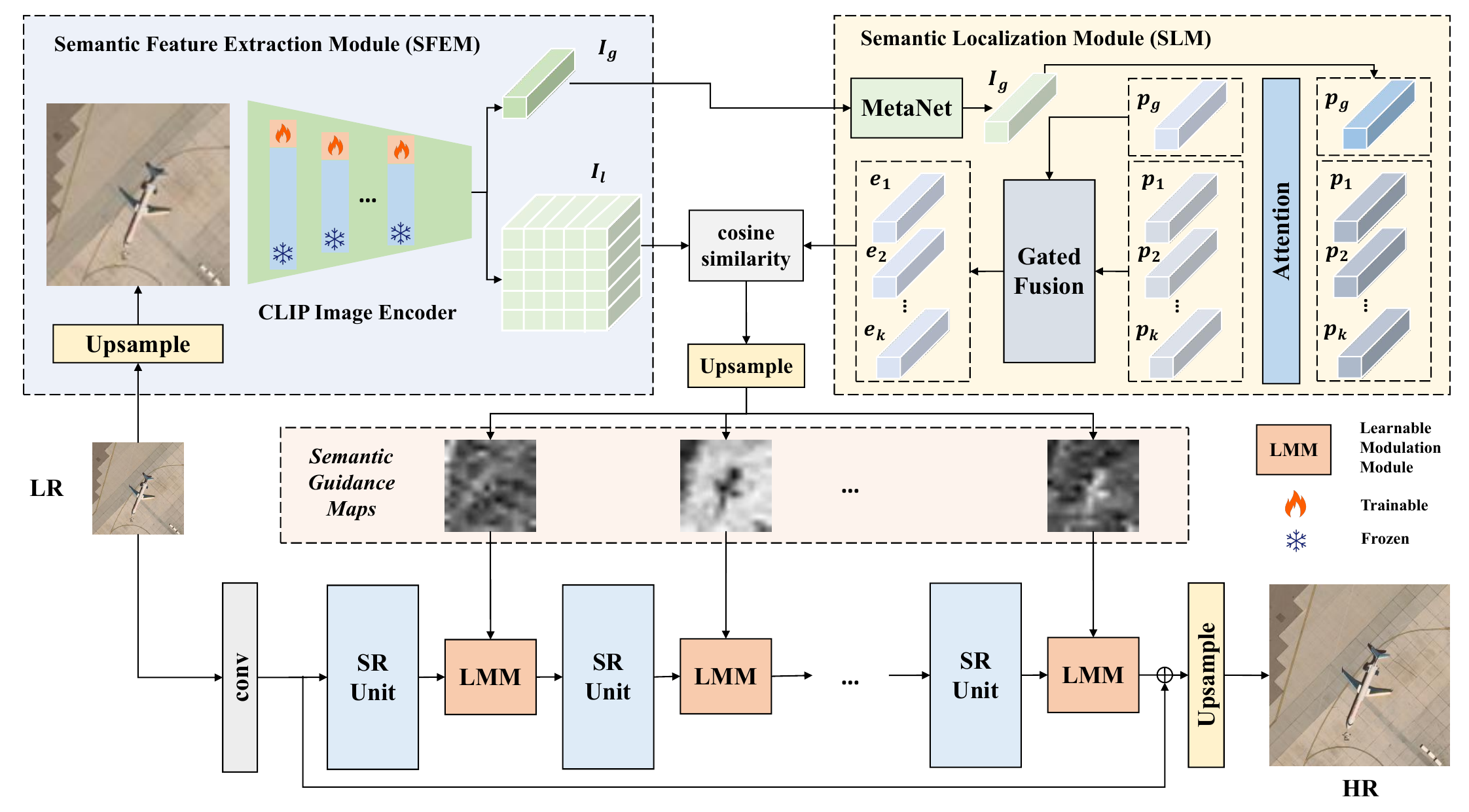}
 	\caption{An overview of the proposed SeG-SR. The LR image is first processed by the Semantic Feature Extraction Module (SFEM) to obtain both global and local semantic features. The global features are fed into the Semantic Localization Module (SLM) to generate per-unit localization embeddings. These embeddings are then matched with the local features to produce semantic guidance maps, which are subsequently used to guide the super-resolution process through Learnable Modulation Module (LMM).
  }
 	\label{fig:flowchart}
\end{figure*}

\color{black} 

\section{Proposed Method} \label{method}

The overall architecture of our proposed SeG-SR framework is illustrated in Fig. \ref{fig:flowchart}. It consists of three core components: the Semantic Feature Extraction Module (SFEM), the Semantic Localization Module (SLM), and the Learnable Modulation Module (LMM). Given LR input, we first apply bilinear upsampling and then feed the result into the image encoder of a CLIP model fine-tuned using Low-Rank Adaptation (LoRA) \cite{hu2022lora} to extract both global and local semantic features. The global semantic features are passed to the SLM, which generates a unique localization embedding for each SR unit. These localization embeddings are then matched with the corresponding local semantic features to produce a set of semantic guidance maps. Finally, these maps are passed through the LMM, which modulates and directs the SR process at each unit level.

\subsection{Semantic Feature Extraction Module}
The Semantic Feature Extraction Module (SFEM) is designed to extract semantic features from the LR input using the image encoder of a fine-tuned CLIP model. Let the LR input be denoted as \(\mathbf{X} \in \mathbb{R}^{H \times W \times 3}\). First, \(\mathbf{X}\) is upsampled via bilinear interpolation to obtain \(\mathbf{X}_{up} \in \mathbb{R}^{H \cdot s \times W \cdot s \times 3}\) , where \(s\) represents the upsampling scale. The upsampled image \(\mathbf{X}_{up}\) is then fed into the fine-tuned CLIP image encoder to extract both global and local semantic features, denoted as \(\mathbf{I}_{g} \in \mathbb{R}^{1 \times C}\) and \(\mathbf{I}_{l} \in \mathbb{R}^{H_c \times W_c \times C}\), respectively. \(H_c\) and \(W_c\) represent the length and width of the feature map. Taking the ViT-B-16 version as an example, \(H_c = H \cdot s / 16\).

Here, \(\mathbf{I}_{g}\) corresponds to the first token output of the CLIP image encoder, commonly referred to as the \([CLS]\) token, which encapsulates the most discriminative semantic information of the entire image. In contrast, \(\mathbf{I}_{l}\) comprises the remaining tokens, each representing semantic information from different spatial locations within the image. This process can be expressed as:
\begin{equation}\label{eq:clip}
    \begin{aligned}
        \mathbf{I}_{g}, \mathbf{I}_{l} = \Phi_{ft-CLIP}(f_{up}(\mathbf{X})),
    \end{aligned}
\end{equation}
where \(\Phi_{ft-CLIP}(\cdot)\) represents the fine-tuned CLIP image encoder and \(f_{up}(\cdot)\) represents the upsampling operation.

The original CLIP model is primarily designed for general scene understanding tasks. However, due to the substantial domain gap between general and remote sensing scenes, directly applying CLIP to remote sensing tasks often results in performance degradation. To address this issue, we adopt Low-Rank Adaptation (LoRA) \cite{hu2022lora} by introducing additional trainable parameters, enabling efficient fine-tuning of the model. This adaptation can be formulated as follows:
\begin{equation}\label{eq:lora}
    \begin{aligned}
        \mathbf{W}^{*} = \mathbf{W} + \Delta \mathbf{W} = \mathbf{W} + \mathbf{A}\mathbf{B}^T,
    \end{aligned}
\end{equation}
where \(\mathbf{W} \in \mathbb{R}^{d \times d}\) represents the frozen pre-trained weight of the CLIP image encoder, while \(\mathbf{A} \in \mathbb{R}^{d \times r}\) and \(\mathbf{B} \in \mathbb{R}^{c \times r}\), are the learnable low-rank matrices introduced for fine-tuning, with \(r << d\).
We apply LoRA fine-tuning to the linear projection layers which generate the self-attention matrices Q, K, V, and the linear layers of the feed-forward networks (FFNs) in the CLIP image encoder. The LoRA process in self-attention can be formally represented as:
\begin{equation}
    \begin{aligned}
        \mathbf{Q} = \mathbf{W}_{q}^{*}\mathbf{F}, \mathbf{K} = \mathbf{W}_{k}^{*}\mathbf{F}, \mathbf{V} = \mathbf{W}_{q}^{*}\mathbf{F}\\
        Attn=Softmax(\frac{\mathbf{Q}\mathbf{K}^{T}}{\sqrt{d}})\mathbf{V},
    \end{aligned}
\end{equation}
where \(\mathbf{F}\) represents the input feature and \(\mathbf{W}_{q}^{*}\), \(\mathbf{W}_{k}^{*}\), \(\mathbf{W}_{v}^{*}\) generated by Eq. \ref{eq:lora}  represent the fine-tuned weight matrices.

The LoRA process in FFNs can be denoted as:
\begin{equation}
    \begin{aligned}
        \mathbf{F}_{1} &= \mathbf{W}_{1}^{*}\mathbf{F}\\
        \mathbf{F}_{2} &= GeLU(\mathbf{W}_{2}^{*}\mathbf{F}),
    \end{aligned}
\end{equation}
where \(\mathbf{W}_{1}^{*}\) and \(\mathbf{W}_{2}^{*}\) represent the fine-tuned weight matrices.

\subsection{Semantic Localization Module}

\begin{figure*}
 	\centering
 	\includegraphics[width=\textwidth]{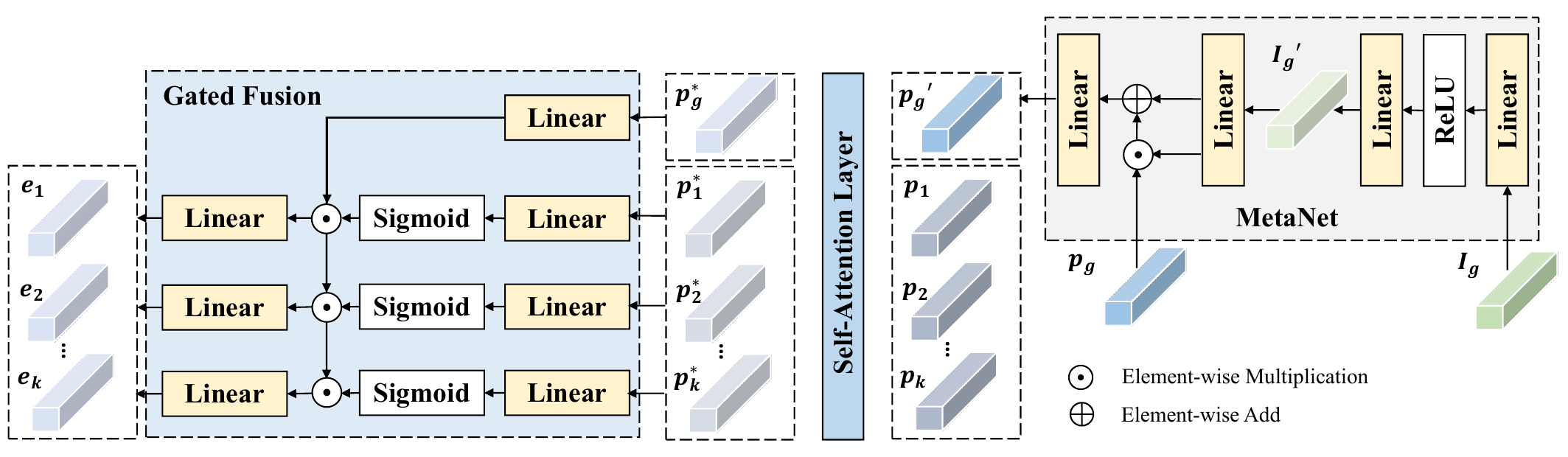}
 	\caption{The structure of the proposed SLM. The MetaNet is used to generate the global feature vector, the self-attention layer interacts with and integrates global and local feature vectors, and the Gated Fusion module produces the final semantic localization embeddings.}
 	\label{fig:slm}
\end{figure*}

The Semantic Localization Module (SLM) is designed to generate a localization embedding for each SR unit, which is similar to the role of CLIP’s Text Encoder. These embeddings are used to match against the local semantic features extracted by the SFEM, ultimately producing the semantic guidance maps. The architecture of SLM is illustrated in Fig. \ref{fig:slm}.

If the number of SR units is \(k\), \(k+1\) feature vectors \(\mathbf{p}\) are initialized first, where \(k\) vectors \([\mathbf{p}_1, \mathbf{p}_2, ..., \mathbf{p}_k], \mathbf{p}_i \in \mathbb{R}^{1 \times C}\) are the local descriptors corresponding to each SR unit and \(\mathbf{p}_g \in \mathbb{R}^{1 \times C}\) serves as a global descriptor.

To incorporate semantic information, \(\mathbf{p}_g\) is fused with the global semantic feature \(\mathbf{I}_{g}\) via a small network called MetaNet. Specifically, \(\mathbf{I}_{g}\) is first passed through a multilayer perceptron (MLP) to get \(\mathbf{I}_{g}^{'}\), followed by a linear projection operation. The result is then element-wise multiplied with \(\mathbf{p}_g\), and the product is added back. Finally, the sum is passed through a linear layer to produce the refined global vector \(\mathbf{p}_g^{'}\).This process can be described as:

\begin{equation}
    \begin{aligned}
        \mathbf{I}_{g}^{'} &= Linear(ReLU(Linear(\mathbf{I}_{g})))\\
        \mathbf{p}_g^{'} &= Linear(Linear(\mathbf{I}_{g}^{'}) \odot \mathbf{p}_g^{'} + Linear(\mathbf{I}_{g}^{'})),
    \end{aligned}
\end{equation}
where \(\odot\) denotes element-wise multiplication. Subsequently, both \(\mathbf{p}_g^{'}\) and \([\mathbf{p}_1, \mathbf{p}_2, ..., \mathbf{p}_k]\) are fed into a self-attention layer for interaction, producing the refined feature vectors \(\mathbf{p}_g^{*}\) and \([\mathbf{p}_1^*, \mathbf{p}_2^*, ..., \mathbf{p}_k^*]\). It can be formulated as follows:
\begin{equation}
    \begin{aligned}
        \mathbf{p}_g^{*}, \mathbf{p}_1^*, \mathbf{p}_2^*, ..., \mathbf{p}_k^*= Attn(\mathbf{p}_g^{'}, \mathbf{p}_1, \mathbf{p}_2, ..., \mathbf{p}_k).
    \end{aligned}
\end{equation}

Next, we use the vector \(\mathbf{p}_i^*\) as a gating mechanism to modulate \(\mathbf{p}_g^{*}\). Specifically, each \(\mathbf{p}_i^*\) is first passed through a linear projection followed by a sigmoid activation to generate a gating vector. This gating vector is then element-wise multiplied with a linearly projected version of \(\mathbf{p}_g^{*}\). The product is further passed through another linear layer to generate the final semantic localization embeddings \(\mathbf{e}_i \in \mathbb{R}^{1 \times C}\) for each SR unit. Formally, the process for each of the \(k\) SR units can be expressed as:
\begin{equation}
    \begin{aligned}
        \mathbf{e}_i = Linear(Sigmoid(Linear(\mathbf{p}_i^*)) \odot Linear(\mathbf{p}_g^*))
    \end{aligned}
\end{equation}

Finally, for the \(i\)-th SR unit, we compute the cosine similarity between \(\mathbf{e}_i\) and the local semantic features \(\mathbf{I}_l\) to obtain the corresponding low-resolution semantic guidance map \(\mathbf{M}_{i}^{lr} \in \mathbb{R}^{H_c \times W_c \times 1}\). This map is then upsampled to generate the semantic guidance map \(\mathbf{M}_{i} \in \mathbb{R}^{H \times W \times 1} \) which matches the size of the input LR image. The process can be expressed as:
\begin{equation}
    \begin{aligned}
        \mathbf{M}_{i}^{lr} &= \frac{\mathbf{e}_i \cdot \mathbf{I}_l}{\Vert \mathbf{e}_i\Vert_2 \cdot \Vert \mathbf{I}_l \Vert_F} \\
        \mathbf{M}_{i} &= Up(\mathbf{M}_{i}^{lr}),
    \end{aligned}
\end{equation}
where \(\Vert \cdot \Vert_2\) represents the L2 norm, and \(\Vert \cdot \Vert_F\) represents the Frobenius norm.

\subsection{Learnable Modulation Module}

\begin{figure}[t]
 	\centering
 	\includegraphics[width=0.48\textwidth]{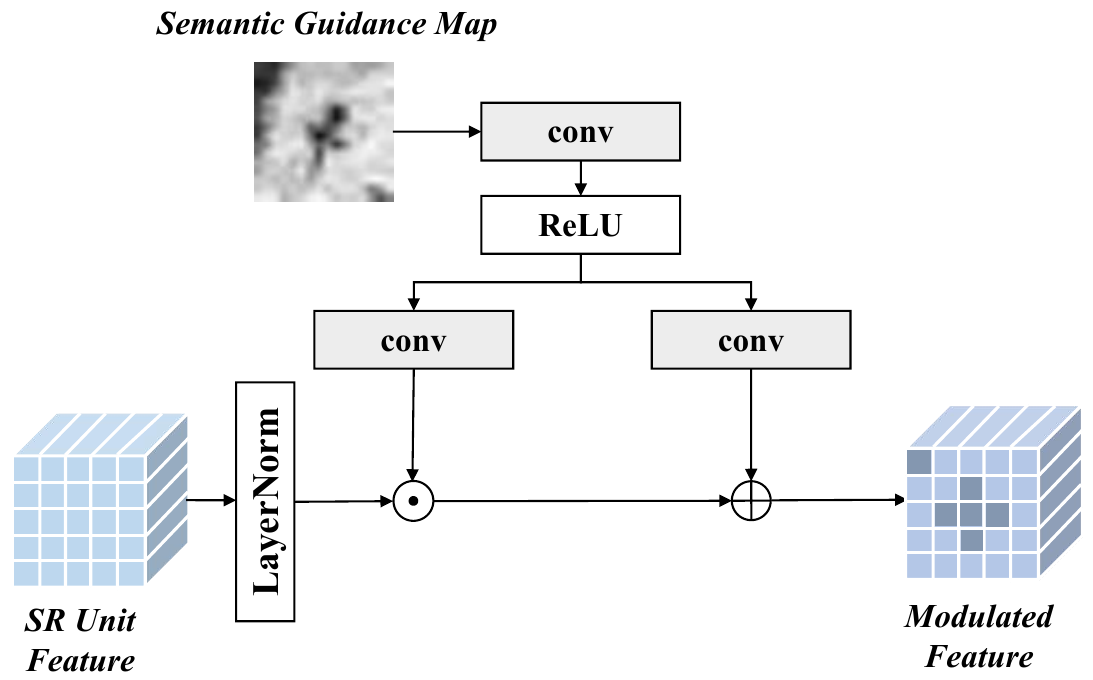}
 	\caption{The structure of the proposed LMM. LMM incorporates semantic information into the SR process by modulating the output features of each SR unit using the corresponding semantic guidance map}
 	\label{fig:lmm}
\end{figure}

The Learnable Modulation Module (LMM) modulates the feature \(\mathbf{F}_i\) output by the \(i\)-th SR unit using the semantic guidance map \(\mathbf{M}_i\), thereby incorporating semantic information into the SR process. The structure of LMM is illustrated in Fig. \ref{fig:lmm}.

Specifically, \(\mathbf{M}_i\) is first processed by a shared convolutional layer followed by a non-linear activation function. The result is then passed through two separate convolutional layers to generate the modulation parameters: gain \(\mathbf{g}_i\) and bias \(\mathbf{b}_i\). Meanwhile, \(\mathbf{F}_i\) is first normalized by layer normalization. Then the normalized features is modulated by element-wise multiplication with the \(\mathbf{g}_i\), followed by an element-wise addition of the \(\mathbf{b}_i\). The process can be expressed as:
\begin{equation}
    \begin{aligned}
        \mathbf{g}_i &= Conv(ReLU(Conv(\mathbf{M}_i)))\\
        \mathbf{b}_i &= Conv(ReLU(Conv(\mathbf{M}_i))) \\
        \mathbf{F}_i^{out} &= \mathbf{g}_i \odot LN(\mathbf{F}_i) + \mathbf{b}_i,
    \end{aligned}
\end{equation}
where \(LN(\cdot)\) represents layer normalization and \(\mathbf{F}_i^{out}\) represents modulated feature of the \(i\)-th SR unit.

\subsection{Super Resolution Network}

We adopt HAT \cite{chen2023activating} as the baseline architecture for our SR network and use Residual Hybrid Attention Group (RHAG) as the SR unit. The LR input \(\mathbf{X}\) is first passed through a convolutional layer to extract shallow features, denoted as \(\mathbf{F}_{shallow}\) . These features are then sequentially processed by a series of SR units.

For each SR unit, the output feature  \(\mathbf{F}_{i}\) is modulated by its corresponding semantic guidance map  \(\mathbf{M}_{i}\) through the Learnable Modulation Module (LMM). After processing through all \(k\) SR units and applying semantic modulation, the resulting output is denoted as  \(\mathbf{F}_{k}^{out}\).

To enhance information flow and preserve low-level details, we employ a residual connection by adding \(\mathbf{F}_{k}^{out}\) to \(\mathbf{F}_{shallow}\). The combined feature is then passed through the final upsampling module, which consists of a convolutional layer followed by a pixel shuffle operation to generate the high-resolution output. The process described above can be formally expressed as:
\begin{equation}
    \begin{aligned}
        \mathbf{F}_{shallow} &= Conv(X) \\
        \mathbf{F}_{i} &= \Phi_{SR-unit}^{i}(\mathbf{F}_{i-1}^{out}) \\
        \mathbf{F}_{i}^{out} &= \Phi_{LMM}^{i}(\mathbf{F}_{i}, \mathbf{M}_{i})\\
        \mathbf{\hat{Y}} &= Conv(Up(\mathbf{F}_{k}^{out} +  \mathbf{F}_{shallow})),
    \end{aligned}
\end{equation}
where \(\mathbf{\hat{Y}} \in \mathbb{R}^{s\cdot H \times s \cdot W \times 3}\) represents reconstructed result. In Section \ref{experiment}, we also explore the use of alternative architectures as SR units within our framework. Detailed configurations and corresponding performance comparisons are provided in that section.

As for training, we train our SeG-SR framework using the L1 loss function, which measures the pixel-wise absolute difference between the reconstructed image and the ground truth:
\begin{equation}
    \begin{aligned}
        L = |\mathbf{\hat{Y}} - \mathbf{Y}|,
    \end{aligned}
\end{equation}
where \(\mathbf{Y} \in \mathbb{R}^{s\cdot H \times s \cdot W \times 3}\) represents HR groundtruth.

\section{Experiment}\label{experiment}

In this section, we present the datasets and experimental settings to validate the performance of our SeG-SR. Additionally, we show the reconstructed results of SeG-SR and conduct ablation studies on it.

\begin{figure*}
    \centering    \includegraphics[width=0.95\textwidth]{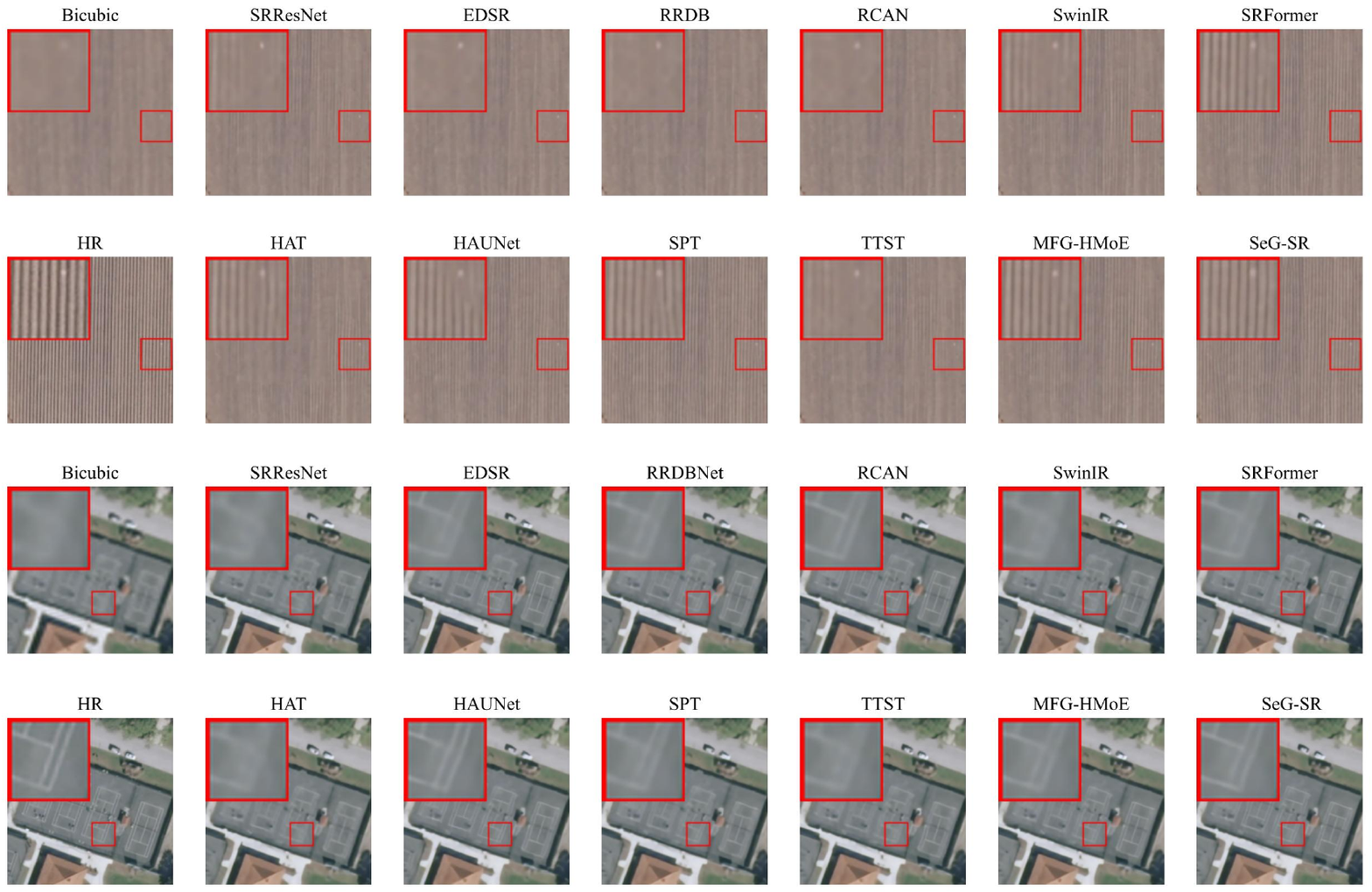}	
    \caption{Super-resolution (\(\times\) 4) results of various SR methods on the UCMerced dataset. The first two rows display reconstruction outputs from image "agricultural 85", while the latter two rows show results from image "tennis court 98".}
	\label{fig:uc vis}
\end{figure*}

\begin{figure*}
    \centering    \includegraphics[width=0.95\textwidth]{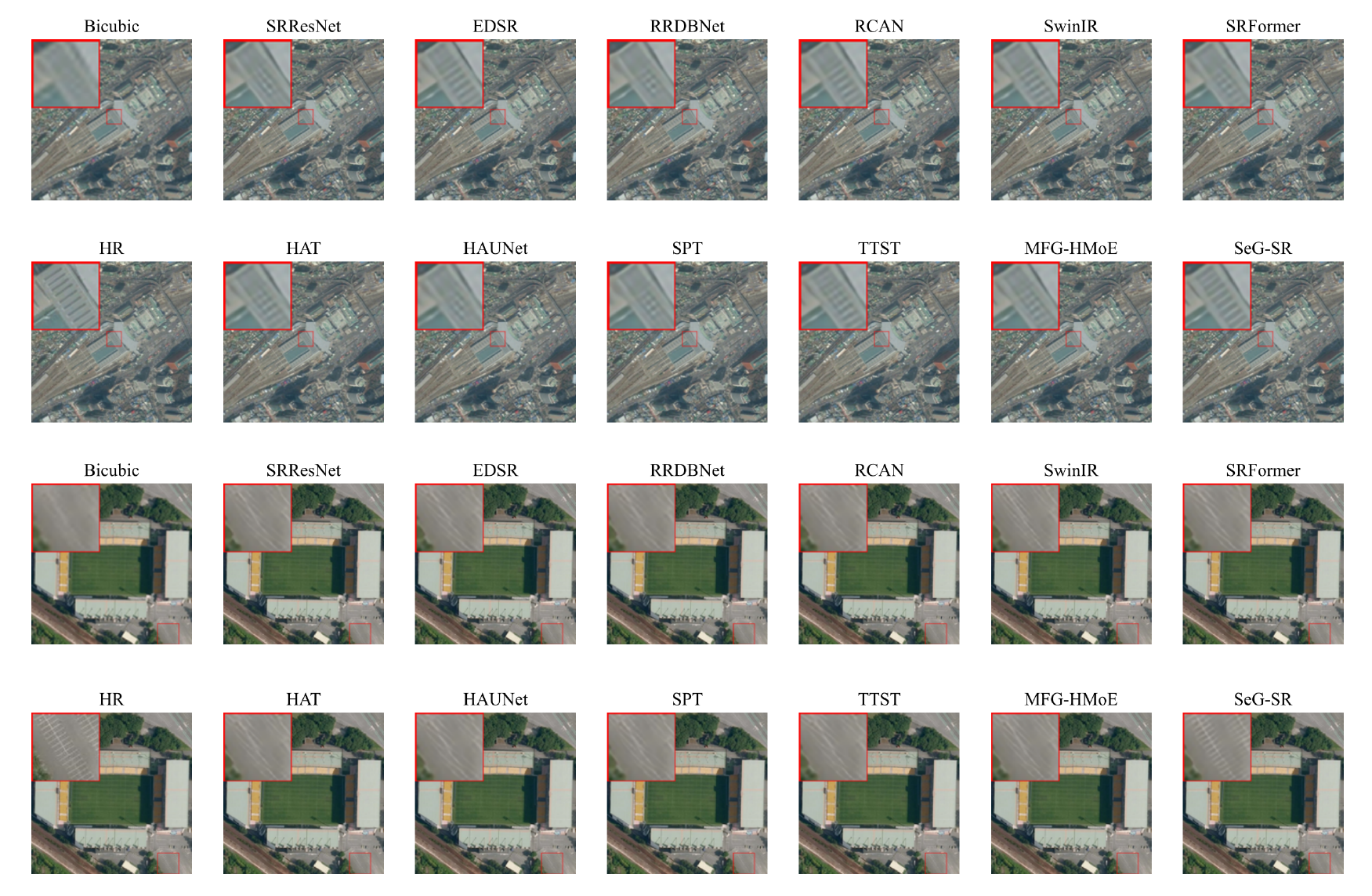}	
    \caption{Super-resolution (\(\times\) 4) results of various SR methods on the AID dataset. The first two rows display reconstruction outputs from image "railway station 127", while the latter two rows show results from image "stadium 192".}
	\label{fig:aid vis}
\end{figure*}

\subsection{Datasets and Experimental Setup}

We evaluate the effectiveness of our proposed SeG-SR on three widely used remote sensing datasets: UCMerced \cite{yang2010bag}, AID \cite{xia2017aid}, and SIRI-WHU \cite{zhao2016fisher, zhao2016dirichlet, zhu2016bag}.  Since our primary focus is on restoring information lost due to reduced resolution, the HR ground truth is taken directly from the original dataset images, while the corresponding LR inputs are generated via bicubic downsampling.
\color{black}

The UCMerced Land Use Dataset consists of images manually extracted from the United States Geological Survey (USGS) National Map Urban Area Imagery collection. These images cover a wide range of urban regions across the United States, spanning 21 scene categories, each containing 100 images, for a total of 2,100 images. Each image is of size \(256 \times 256\) pixels, with a spatial resolution of 0.3 meters per pixel. We randomly split the images in each category into training and testing sets at a 3:1 ratio, resulting in 1,575 training images and 525 testing images.

The AID (Aerial Image Dataset) is a large-scale benchmark dataset designed for aerial scene classification. It contains 10,000 images collected from Google Earth, covering 30 different scene categories across various countries and regions worldwide. Each image has a size of \(600 \times 600\) pixels with a spatial resolution of 0.5 meters per pixel. The number of images per category ranges from approximately 200 to 400. We randomly divide the dataset into training and testing sets using a 4:1 ratio, yielding 8,000 training images and 2,000 testing images.

The SIRI-WHU dataset was collected from Google Earth and primarily covers urban areas in China. It consists of 12 categories, each containing 200 images with a spatial resolution of 2 meters and a size of \(200 \times 200 \)pixels. We randomly divided the images in each category into training and testing subsets with a 3:1 ratio, resulting in 1800 training images and 600 testing images.
\color{black}

Our proposed SeG-SR method was compared with 11 state-of-the-art SR methods from both general and remote sensing domains, including CNN-based methods (SRResNet \cite{ledig2017photo}, EDSR \cite{lim2017enhanced}, RRDBNet \cite{wang2018esrgan}), attention-based methods (RCAN \cite{zhang2018image}, HAUNet \cite{wang2023hybrid}), and Transformer-based methods (SwinIR \cite{liang2021swinir}, SRFormer \cite{zhou2023srformer}, HAT \cite{chen2023activating}, TTST \cite{xiao2024ttst}, SPT \cite{hao2024scale}, MFG-HMoE \cite{chen2025heterogeneous}). SRResNet designed a feature extraction network with residual connections. EDSR removed BN layers from SRResNet while aggregating multi-level feature information. RRDBNet combined dense connections with residual onnections for feature extraction. RCAN introduced channel attention to enhance feature representation. HAUNet, built upon a U-Net \cite{ronneberger2015u} architecture, employs convolutional attention for single-scale feature extraction and multi-head cross-attention for cross-scale information integration. SwinIR adapted Swin Transformer for SR tasks. SRFormer utilizes Permuted Self-Attention to expand the attention window. HAT integrates hybrid attention mechanisms to further unleash Transformer's potential. TTST improves computational efficiency through a token selection mechanism that eliminates redundant information interference. SPT combines back-projection mechanisms with Transformer architecture. MFG-HMoE employs MoE to enable pixel-adaptive reconstruction strategy selection. 

To ensure a fair comparison across all methods, we adopt a consistent training protocol. Specifically, all models are trained using LR input images with a fixed resolution of \(64 \times 64\) pixels. Data augmentation techniques, including flipping and rotation, are applied uniformly across all methods. The batch size is set to 4 for all experiments. Optimization-related parameters—such as optimizer type, learning rate, and learning rate decay strategy—are configured according to the optimal settings reported in each corresponding paper. All methods are thoroughly trained and tuned, and the best-performing model on the test set is selected for comparison. All experiments are conducted on a system equipped with an Intel(R) Xeon(R) Gold 6330 CPU @ 2.00GHz and an NVIDIA GeForce RTX 4090 GPU, running Ubuntu OS version 20.04.6.

\subsection{Evaluation Metrics}

Our purpose is to restore details in remote sensing images that are lost due to resolution degradation, rather than to hallucinate visually appealing but potentially spurious details. Such details may not reflect the actual ground truth \cite{el2022bigprior} and could even impair the performance of downstream applications. Based on this objective, we selected pixel-level fidelity metrics \cite{sun2022rethinking, zhang2020unified}, Peak Signal-to-Noise Ratio (PSNR) and Structural Similarity Index Measure (SSIM) \cite{wang2004image}, as our primary evaluation criteria.

\color{black}

PSNR quantifies the distortion between a reconstructed image \(\mathbf{\hat{Y}}\) and an HR ground truth \(\mathbf{Y}\) by computing the Mean Squared Error (MSE) and then converting it into a logarithmic scale that reflects the ratio between the maximum possible signal power and the noise power. It is defined as:

\begin{equation}
    \begin{aligned}
        \mathrm{PSNR} = 10 \cdot \log_{10}(\frac{MAX^2}{\Vert \mathbf{\hat{Y}} - \mathbf{Y}\Vert_2^2})
    \end{aligned}
\end{equation}
where \(MAX^2\) denotes the maximum possible pixel value (typically 255 for 8-bit image) and higher PSNR values indicate better reconstruction quality.

SSIM evaluates image similarity based on three components: luminance, contrast, and structure. SSIM between a reconstructed image \(\mathbf{\hat{Y}}\) and an HR ground truth \(\mathbf{Y}\) is computed as:
\begin{equation}
\mathrm{SSIM}(\mathbf{\hat{Y}},\mathbf{Y})=\frac{(2\mu_{\mathbf{\hat{Y}}}\mu_{\mathbf{Y}}+C_1)(2\sigma_{\mathbf{\hat{Y}}{\mathbf{Y}}}+C_2)}{(\mu_{\mathbf{\hat{Y}}}^2+\mu_{\mathbf{Y}}^2+C_1)(\sigma_{\mathbf{\hat{Y}}}^2+\sigma_{\mathbf{Y}}^2+C_2)}
\end{equation}
where \(\mu\) represents the mean value, \(\sigma\) represents the variance or covariance, and constants \(C_1\) and \(C_2\) are included to stabilize the division in cases of weak denominators. Higher SSIM values indicate better reconstruction quality.

However, PSNR and SSIM have limitations—for instance, they may assign high scores to excessively smoothed regions. To achieve a more comprehensive evaluation, we also incorporated perceptual quality assessments, such as Learned Perceptual Image Patch Similarity (LPIPS) \cite{zhang2018unreasonable} and CLIPSCORE \cite{wolters2023zooming}, as auxiliary metrics.

LPIPS is a deep feature–based perceptual similarity measure, which compares intermediate representations extracted from a pretrained convolutional neural network across multiple layers. Its formulation is as follows:

\begin{equation}
\mathbf{LPIPS}(\mathbf{Y},\mathbf{\hat{Y}})=\sum_l\frac{1}{H_lW_l}\sum_{h,w}||\mathbf{w}_l\odot(\mathbf{f}_{hw}^l-\mathbf{\hat{f}}_{hw}^l)||_2^2,
\end{equation}
where \(H_l\) and \(H_l\) represents the height and width of the feature map at the \(l\)-th layer, \(\mathbf{w}_l\) represents the channel weighting vector of the \(l\)-th layer, and \(\mathbf{f}_{hw}^l\) and \(\mathbf{\hat{f}}_{hw}^l\) represent the normalized feature maps of the reconstructed image and the HR ground truth at the \(l\)-th layer of the network. Smaller LPIPS values indicate better perceptual quality.

CLIPSCORE measures the distance between ground-truth and super-resolution results in the embedding space of the CLIP model, serving as a reference-based perceptual quality metric aligned with human perception. Its formulation is as follows:
\begin{equation}
\mathbf{CLIPSCORE}(\mathbf{{Y}},\mathbf{\hat{Y}})= cos(\Phi_{CLIP}(\mathbf{{Y}}), \Phi_{CLIP}(\mathbf{\hat{Y}})),
\end{equation}
where \(cos(\cdot, \cdot)\) represents the cosine similarity, \(\Phi_{CLIP}(\cdot)\) represents the CLIP model. Higher CLIPSCORE values indicate better perceptual quality.

\color{black}

\subsection{Implementation Details of SeG-SR}

\subsubsection{Parameter Settings}

In terms of network architecture, we follow the design of HAT, where each SR unit consists of a RHAG, with the number of stacked SR units set to \(k=6\). For the SFEM, the rank \(r\) of the low-rank matrices that can be learned for fine-tuning is configured as 32.

\subsubsection{Training details}

For both datasets, we employ the Adam optimizer ~\cite{kingma-adam2014-76} for training, with an initial learning rate set to $10^{-4}$.

For the UCMerced dataset and the SIRI-WHU dataset, training is conducted for a total of 80,000 iterations, and the learning rate is reduced by half at the 50,000th iteration to facilitate convergence. For the AID dataset, training proceeds for 120,000 iterations, with the learning rate halved at both the 60,000th and 100,000th iterations. 
\color{black}

\begin{table}[hbt]
\centering
\caption{Pixel-level fidelity comparison of different methods on UCMerced dataset. {\color{red}Red} indicates the best results, while {\color{blue}Blue} represents the second-best results.}
\scalebox{1.00}{
\begin{tabular}{lcccc}
\toprule
\multirow{2}{*}{Method} & \multicolumn{2}{c}{$\times 2$} & \multicolumn{2}{c}{$\times 4$} \\
\cmidrule(lr){2-3} \cmidrule(lr){4-5}
 & SSIM & {\color{black}PSNR (dB)} & SSIM & {\color{black}PSNR (dB)} \\
\midrule
SRResNet \cite{ledig2017photo} & 0.9367 & 35.3248 & 0.7802 & 28.8762 \\
EDSR-L \cite{lim2017enhanced} & 0.9383 & 35.5104 & 0.7840 & 28.9274 \\
RRDBNet \cite{wang2018esrgan} & 0.9335 & 35.0327 & 0.7839 & 28.9929 \\
RCAN \cite{zhang2018image} & 0.9386 & 35.5456 & 0.7860 & 29.0668 \\
SwinIR \cite{liang2021swinir} & 0.9390 & 35.5463 & 0.7877 & 29.0573 \\
SRFormer \cite{zhou2023srformer} & 0.9391 & 35.5784 & 0.7888 & 29.1128 \\
HAT \cite{chen2023activating} & 0.9398 & 35.6829 & 0.7893 & 29.0985 \\
HAUNet \cite{wang2023hybrid} & 0.9389 & 35.6016 & 0.7894 & 29.0804 \\
SPT \cite{hao2024scale} & {0.9394} & {35.6318} & {0.7889} & {29.1108} \\
TTST \cite{xiao2024ttst} & 0.9400 & 35.7058 & 0.7895 & 29.1405 \\
MFG-HMoE \cite{chen2025heterogeneous} & {\color{blue}0.9409} & {\color{blue}35.8110} & {\color{blue}0.7954} & {\color{blue}29.2882} \\
SeG-SR (Ours) & {\color{red}0.9415} & {\color{red}35.8757} & {\color{red}0.7961} & {\color{red}29.3042} \\
\bottomrule
\end{tabular}
}
\label{tab:compare}
\end{table}

\begin{table}[hbt]
\centering
\caption{Comparison of different methods on AID dataset. {\color{red}Red} indicates the best results, while {\color{blue}Blue} represents the second-best results.}
\scalebox{1.00}{
\begin{tabular}{lcccc}
\toprule
\multirow{2}{*}{Method} & \multicolumn{2}{c}{$\times 2$} & \multicolumn{2}{c}{$\times 4$} \\
\cmidrule(lr){2-3} \cmidrule(lr){4-5}
 & SSIM & {\color{black}PSNR (dB)} & SSIM & {\color{black}PSNR (dB)} \\
\midrule
SRResNet \cite{ledig2017photo} & 0.9427 & 36.3866 & 0.8044 & 30.4003 \\
EDSR-L \cite{lim2017enhanced} & 0.9438 & 36.4926 & 0.8091 & 30.5434 \\
RRDBNet \cite{wang2018esrgan} & 0.9407 & 36.2417 & 0.8063 & 30.4646 \\
RCAN \cite{zhang2018image} & 0.9441 & 36.5261 & 0.8098 & 30.5880 \\
SwinIR \cite{liang2021swinir} & 0.9434 & 36.4428 & 0.8089 & 30.5357 \\
SRFormer \cite{zhou2023srformer} & 0.9439 & 36.5017 & 0.8091 & 30.5562 \\
HAT \cite{chen2023activating} & 0.9439 & 36.4975 & 0.8104 & 30.5920 \\
HAUNet \cite{wang2023hybrid} & 0.9436 & 36.4883 & 0.8079 & 30.5571 \\
{SPT \cite{hao2024scale}} & {0.9431} & {36.4542} & {0.8094} & {30.5847} \\
TTST \cite{xiao2024ttst} & 0.9441 & 36.5142 & 0.8108 & 30.6108 \\
MFG-HMoE \cite{chen2025heterogeneous} & {\color{blue}0.9443} & {\color{red}36.5399} & {\color{blue}0.8110} & {\color{blue}30.6167} \\
SeG-SR (Ours) & {\color{red}0.9444} & {\color{blue}36.5316} & {\color{red}0.8110} & {\color{red}30.6533} \\
\bottomrule
\end{tabular}
}
\label{tab:compare2}
\end{table}

\begin{table}[hbt]
\centering
\caption{Pixel-level fidelity comparison of different methods on SIRI-WHU dataset. {\color{red}Red} indicates the best results, while {\color{blue}Blue} represents the second-best results.}
\scalebox{1.00}{
\begin{tabular}{lcccc}
\toprule
\multirow{2}{*}{Method} & \multicolumn{2}{c}{$\times 2$} & \multicolumn{2}{c}{$\times 4$} \\
\cmidrule(lr){2-3} \cmidrule(lr){4-5}
 & SSIM & {\color{black}PSNR (dB)} & SSIM & {\color{black}PSNR (dB)} \\
\midrule
SRResNet \cite{ledig2017photo} & 0.9413 & 36.3453 & 0.8071 & 30.8895 \\
EDSR-L \cite{lim2017enhanced} & 0.9412 & 36.3477 & 0.8078 & 30.9052 \\
RRDBNet \cite{wang2018esrgan} & 0.9397 & 36.2309 & 0.8084 & 30.9288 \\
RCAN \cite{zhang2018image} & 0.9421 & 36.4012 & 0.8088 & 30.9456 \\
SwinIR \cite{liang2021swinir} & 0.9416 & 36.3858 & 0.8081 & 30.9301 \\
SRFormer \cite{zhou2023srformer} & 0.9421 & 36.4269 & 0.8098 & 30.9488 \\
HAT \cite{chen2023activating} & 0.9420 & 36.4186 & 0.8087 & 30.9299 \\
HAUNet \cite{wang2023hybrid} & 0.9419 & 36.4207 & 0.8085 & 30.9502 \\
SPT \cite{hao2024scale} & 0.9421 & 36.4259 & 0.8092 & 30.9433 \\
TTST \cite{xiao2024ttst} & {\color{blue}0.9421} & {\color{blue}36.4304} & 0.8088 & 30.9320 \\
MFG-HMoE \cite{chen2025heterogeneous} & 0.9421 & 36.4285 & {\color{red}0.8099} & {\color{blue}30.9617} \\
SeG-SR (Ours) & {\color{red}0.9422} & {\color{red}36.4418} & {\color{blue}0.8098} & {\color{red}30.9702} \\
\bottomrule
\end{tabular}
}
\label{tab:compare3}
\end{table}

\begin{table}[hbt]
\centering
\caption{Perceptual quality of different methods on three datasets in $\times 4$ SR task. {\color{red}Red} indicates the best results, while {\color{blue}Blue} represents the second-best results.}
\scalebox{0.75}{
\begin{tabular}{lcccccc}
\toprule
\multirow{2}{*}{Method} & \multicolumn{2}{c}{UCMerced} & \multicolumn{2}{c}{AID} & \multicolumn{2}{c}{SIRI} \\
\cmidrule(lr){2-3} \cmidrule(lr){4-5} \cmidrule(lr){6-7}
& LPIPS$\downarrow$ & CLIPScore$\uparrow$ & LPIPS$\downarrow$ & CLIPScore$\uparrow$ & LPIPS$\downarrow$ & CLIPScore$\uparrow$ \\
\midrule
SRResNet \cite{ledig2017photo}   & 0.2962 & 0.8850 & 0.3566 & 0.9334 & 0.3451 & 0.8845 \\
EDSR-L \cite{lim2017enhanced}      & 0.2849 & 0.8982 & \textcolor{blue}{0.3514} & 0.9355 & 0.3507 & 0.8882 \\
RRDBNet \cite{wang2018esrgan}       & 0.2909 & 0.8896 & 0.3573 & 0.9345 & 0.3467 & 0.8886 \\
RCAN \cite{zhang2018image}       & 0.2934 & 0.8899 & 0.3564 & 0.9343 & 0.3471 & 0.8886 \\
SwinIR \cite{liang2021swinir}     & 0.2765 & \textcolor{blue}{0.8990} & 0.3544 & 0.9351 & 0.3480 & 0.8877 \\
SRFormer \cite{zhou2023srformer}   & 0.2743 & 0.8967 & 0.3531 & \textcolor{blue}{0.9355} & 0.3463 & 0.8872 \\
HAT \cite{chen2023activating}       & 0.2740 & 0.8981 & 0.3529 & 0.9344 & 0.3453 & 0.8864 \\
HAUNet \cite{wang2023hybrid}     & 0.2774 & 0.8951 & 0.3663 & 0.9343 & 0.3454 & 0.8870 \\
SPT \cite{hao2024scale}        & 0.2725 & 0.8978 & 0.3556 & 0.9349 & 0.3441 & \textcolor{blue}{0.8895} \\
TTST \cite{xiao2024ttst}       & 0.2748 & 0.8976 & 0.3515 & 0.9352 & 0.3454 & 0.8884 \\
MFG-HMoE \cite{chen2025heterogeneous}   & 0.2713 & 0.8982 & 0.3522 & 0.9352 & \textcolor{blue}{0.3429} & 0.8889 \\
SeG-SR (Ours)       & \textcolor{red}{0.2704} & \textcolor{red}{0.8991} & \textcolor{red}{0.3489} & \textcolor{red}{0.9356} & \textcolor{red}{0.3430} & \textcolor{red}{0.8893} \\
\hline
\end{tabular}
}
\label{tab:perceptual}
\end{table}

\begin{table*}[hbt]
\centering
\caption{Performance comparison of SeG-SR versus other methods across different scene categories in the UCMerced dataset. {\color{red}Red} indicates the best results, while {\color{blue}Blue} represents the second-best results. The last two columns ($\Delta$ SeG-SR) show the gain of SeG-SR over the baseline method (HAT).}
\scalebox{0.88}{
\begin{tabular}{lcccccccccccccc}
\toprule
\multirow{2}{*}{Class}
  & \multicolumn{2}{c}{HAT}
  & \multicolumn{2}{c}{HAUNet}
  & \multicolumn{2}{c}{SPT}
  & \multicolumn{2}{c}{TTST}
  & \multicolumn{2}{c}{MFG-HMoE}
  & \multicolumn{2}{c}{SeG-SR} 
  & \multicolumn{2}{c}{$\Delta$ SeG-SR} \\
\cmidrule(lr){2-3} \cmidrule(lr){4-5}
\cmidrule(lr){6-7} \cmidrule(lr){8-9}
\cmidrule(lr){10-11} \cmidrule(lr){12-13} \cmidrule(lr){14-15}
  & SSIM & {\color{black}PSNR}
  & SSIM & {\color{black}PSNR}
  & SSIM & {\color{black}PSNR}
  & SSIM & {\color{black}PSNR}
  & SSIM & {\color{black}PSNR}
  & SSIM & {\color{black}PSNR}
  & $\Delta$SSIM & $\Delta$PSNR \\
\midrule
agricultural       & 0.4454 & 25.9089 & 0.4629 & 25.8604 & 0.4237 & 25.4017 & 0.4331 & 25.7061 & \color{blue}0.4795 & \color{red}26.1271 & \color{red}0.4812 & \color{blue}25.9110 & 0.0358 & 0.0021 \\
airplane           & 0.8620 & 30.2158 & 0.8608 & 30.1278 & 0.8621 & 30.1685 & 0.8620 & 30.2693 & \color{blue}0.8640 & \color{blue}30.3049 & \color{red}0.8655 & \color{red}30.3747 & 0.0035 & 0.1589 \\
baseball diamond   & 0.8825 & 35.1967 & 0.8825 & 35.1434 & 0.8821 & 35.0925 & 0.8824 & 35.1594 & \color{blue}0.8834 & \color{red}35.2583 & \color{red}0.8835 & \color{blue}35.2488 & 0.0010 & 0.0521 \\
beach              & 0.9260 & 38.8892 & 0.9261 & 38.8142 & 0.9258 & 38.7887 & 0.9259 & 38.8446 & \color{blue}0.9265 & \color{red}38.9307 & \color{red}0.9266 & \color{blue}38.9262 & 0.0006 & 0.0370 \\
buildings          & 0.7825 & 25.6807 & 0.7811 & 25.6987 & 0.7821 & 25.7315 & 0.7850 & 25.8140 & \color{blue}0.7895 & \color{blue}25.9649 & \color{red}0.7901 & \color{red}26.0421 & 0.0076 & 0.3614 \\
chaparral          & 0.7460 & 27.1512 & 0.7465 & 27.1771 & 0.7462 & 27.1120 & 0.7462 & 27.1656 & \color{blue}0.7468 & \color{blue}27.1811 & \color{red}0.7472 & \color{red}27.1879 & 0.0012 & 0.0367 \\
dense residential  & 0.8254 & 28.1188 & 0.8273 & 28.2054 & 0.8262 & 28.1716 & 0.8272 & 28.2367 & \color{blue}0.8322 & \color{blue}28.4180 & \color{red}0.8325 & \color{red}28.4389 & 0.0071 & 0.3201 \\
forest             & 0.6424 & 27.3370 & \color{blue}0.6432 & \color{red}27.3740 & 0.6429 & 27.3563 & 0.6425 & 27.3477 & 0.6431 & 27.3658 & \color{red}0.6433 & \color{blue}27.3679 & 0.0009 & 0.0309 \\
freeway            & 0.7875 & 29.7325 & 0.7881 & 29.6470 & 0.7882 & 29.7850 & 0.7880 & 29.7476 & \color{blue}0.7936 & \color{blue}29.9376 & \color{red}0.7951 & \color{red}30.0141 & 0.0076 & 0.2816 \\
golf course        & 0.8919 & 35.9789 & 0.8917 & 35.9596 & 0.8919 & 35.9823 & 0.8919 & 35.9965 & \color{blue}0.8922 & \color{red}36.0236 & \color{red}0.8924 & \color{blue}36.0129 & 0.0005 & 0.0340 \\
harbor             & 0.8586 & 23.4707 & 0.8545 & 23.3941 & 0.8604 & 23.5984 & 0.8589 & 23.5816 & \color{red}0.8660 & \color{blue}23.7931 & \color{blue}0.8647 & \color{red}23.8291 & 0.0061 & 0.3584 \\
intersection       & 0.7867 & 27.5717 & 0.7843 & 27.5787 & 0.7891 & 27.6817 & 0.7892 & 27.6748 & \color{red}0.7972 & \color{blue}27.8651 & \color{blue}0.7963 & \color{red}27.8652 & 0.0096 & 0.2935 \\
medium residential & 0.8282 & 29.8682 & 0.8304 & 30.0079 & 0.8318 & 30.0357 & 0.8305 & 29.9994 & \color{blue}0.8331 & \color{blue}30.1281 & \color{red}0.8334 & \color{red}30.1508 & 0.0052 & 0.2826 \\
mobilehomepark     & 0.6349 & 19.4595 & 0.6313 & 19.4214 & 0.6390 & 19.5908 & 0.6397 & 19.6142 & \color{blue}0.6432 & \color{blue}19.6781 & \color{red}0.6461 & \color{red}19.7337 & 0.0112 & 0.2742 \\
overpass           & 0.7395 & 26.8341 & 0.7400 & 26.8618 & 0.7428 & 26.9343 & 0.7388 & 26.7961 & \color{blue}0.7467 & \color{blue}27.0027 & \color{red}0.7487 & \color{red}27.0949 & 0.0092 & 0.2608 \\
parking lot        & 0.8099 & 23.8670 & 0.8012 & 23.5201 & 0.8101 & 23.9266 & 0.8108 & 23.9978 & \color{blue}0.8251 & \color{blue}24.5074 & \color{red}0.8294 & \color{red}24.6582 & 0.0195 & 0.7912 \\
river              & 0.8091 & 30.9202 & \color{blue}0.8110 & 31.0100 & 0.8090 & 30.9528 & 0.8104 & 31.0175 & 0.8109 & \color{blue}31.0350 & \color{red}0.8114 & \color{red}31.0398 & 0.0023 & 0.1196 \\
runway             & 0.8335 & 32.8190 & 0.8316 & 32.7886 & 0.8330 & 32.8235 & 0.8326 & 32.8215 & \color{red}0.8356 & \color{red}33.0137 & \color{blue}0.8355 & \color{blue}32.9687 & 0.0020 & 0.1497 \\
sparse residential & 0.7889 & 29.1434 & 0.7908 & 29.2656 & 0.7890 & 29.2100 & 0.7904 & 29.2274 & \color{red}0.7928 & \color{red}29.3287 & \color{blue}0.7925 & \color{blue}29.2976 & 0.0036 & 0.1542 \\
storage tanks      & 0.8541 & 31.5908 & 0.8534 & 31.5755 & 0.8552 & 31.6992 & 0.8532 & 31.5864 & \color{blue}0.8568 & \color{red}31.7194 & \color{red}0.8572 & \color{blue}31.6928 & 0.0031 & 0.1020 \\
tennis court       & 0.8410 & 31.3149 & 0.8382 & 31.2574 & 0.8382 & 31.2843 & 0.8412 & 31.3464 & \color{blue}0.8446 & \color{blue}31.4679 & \color{red}0.8458 & \color{red}31.5321 & 0.0048 & 0.2172 \\
\midrule
Average            & 0.7893 & 29.0985 & 0.7894 & 29.0804 & 0.7889 & 29.1108 & 0.7895 & 29.1405 & {\color{blue}0.7954} & {\color{blue}29.2882} & {\color{red}0.7961} & {\color{red}29.3042} & 0.0068 & 0.2056 \\
\bottomrule
\end{tabular}
}
\label{tab:ucmerced_comparison}
\end{table*}

\subsection{Comparison with Other Methods}

\subsubsection{Quantitative Analysis}
Tables \ref{tab:compare} and \ref{tab:compare2} present pixel-level fidelity comparisons between our proposed SeG-SR and other state-of-the-art methods on the UCMerced and AID datasets. On UCMerced dataset, SeG-SR achieves the best PSNR and SSIM scores in both \(\times\)2 and \(\times\)4 SR tasks. Notably, in the \(\times\)4 SR setting, SeG-SR outperforms the baseline method HAT by more than 0.2 dB in PSNR, demonstrating a significant improvement.

On the AID dataset, SeG-SR also achieves the best performance for \(\times\)4 SR in both evaluation metrics. For the \(\times\)2 SR task, SeG-SR obtains the highest SSIM score and ranks second in PSNR, with a performance gap of only about 0.02\% compared to the best-performing method, MFG-HMoE. These results highlight the strong competitiveness of SeG-SR in RSISR tasks.

To more thoroughly assess the method’s robustness across different spatial resolutions, we further conducted experiments on the SIRI-WHU dataset, which has a spatial resolution of 2 m. The results on this dataset are reported in Table \ref{tab:compare3}. SeG-SR achieves the best performance in both PSNR and SSIM in the \(\times\)2 SR task. In the \(\times\)4 SR task, SeG-SR attains the highest PSNR and the second-best SSIM, with only a negligible margin of 0.0001 from the best SSIM score. These results, combined with those from the UCMerced and AID datasets, demonstrate the strong generalizability of SeG-SR across different spatial resolutions.

To evaluate visual perceptual quality, we calculate LPIPS and CLIPSCORE on \(\times\)4 super-resolution results across the above three datasets. The results are reported in Table \ref{tab:perceptual}. SeG-SR achieved the best performance on both metrics for the UCMerced and AID datasets. Although it ranked second on the SIRI-WHU dataset, the differences from the best-performing method were marginal (only 0.0001 in LPIPS and 0.0002 in CLIPSCORE). These results highlight the superior perceptual quality achieved by our approach compared to competing methods, demonstrating its ability to recover more faithful and realistic details.

To further analyze the performance per class of different SR methods, Table \ref{tab:ucmerced_comparison} reports the ×4 SR results on individual scene classes in the UCMerced dataset. We conducted experiments using the six best-performing models for the ×4 super-resolution task on the UCMerced dataset. Among the 20 scene classes, our method achieves the best score in at least one evaluation metric in 18 classes. In the remaining two classes, it ranked second for both PSNR and SSIM. This demonstrates the robust adaptability of our method across diverse types of remote sensing scene.
Additionally, an analysis of per-class gains shows that SeG-SR improves PSNR and SSIM in every category relative to the baseline, which further corroborates the method’s effectiveness. Notably, in object-dense, detail-rich classes such as parking lot, harbor, and dense residential, SeG-SR delivered substantial PSNR improvements of 0.7912 dB, 0.3584 dB, and 0.3201 dB, respectively—indicating that the incorporation of semantic information helps the model better interpret and reconstruct complex scenes. 

\color{black}

\subsubsection{Qualitative Analysis}
Fig. \ref{fig:uc vis} and Fig. \ref{fig:aid vis} provide visualization comparisons of the \(\times\)4 SR results produced by different SR methods on the UCMerced and AID datasets, respectively.

For UCMerced dataset, in the “agricultural 85” scene, methods such as EDSR, RRDBNet, and RCAN fail to reconstruct the line structures inside the red square. SRResNet, SwinIR, and TTST recover partial line patterns, while SRFormer, HAT, HAUNet, and SPT can reconstruct most of the lines but with blurring or distortion. MFG-HMoE is able to restore the structure almost completely, but the result lacks sharpness. In contrast, only our SeG-SR clearly reconstructs the line patterns within the red square, delivering superior sharpness and structural fidelity. Also, in the “tennis court 98” scene, SeG-SR also produces the most visually accurate and detailed reconstruction of court lines compared to all other methods.

For AID dataset, in the “railway station 127” scene, SRResNet, RRDBNet, SRFormer, HAT, HAUNet, SPT, TTST, and MFG-HMoE introduce checkerboard artifacts in the building area. EDSR, RCAN, and SwinIR exhibit blurred textures. In contrast, SeG-SR delivers artifact-free and sharply defined results, indicating better visual quality and structural fidelity. In the “stadium 192” scene, nearly all comparison methods fail to accurately reconstruct the parking lines in the red square, producing unrealistic artifacts. In contrast, SeG-SR successfully recovers the parking line details with both clarity and geometric accuracy.

\subsection {Ablation Studies}

In this section, we further validate the effectiveness of the key components in the SeG-SR framework as well as its generalizability. All validations are conducted exclusively on the \(\times\)4 SR task using the UCMerced dataset. We begin by evaluating the individual contributions of the three core modules: SFEM, SLM and LMM. Then we conduct a series of ablation studies on the internal design choices within these modules to identify the optimal configurations. Finally, to demonstrate the general applicability of our approach, we transfer SeG-SR to other super-resolution architectures based on stacked SR units. We reconfigure these baseline methods using our framework and empirically confirm its adaptability and compatibility across diverse model structures.

\subsubsection{Effectiveness of the key modules}

We begin by conducting an ablation study to evaluate the effectiveness of the three key modules in our framework. The results are summarized in Table \ref{tab:ablation}. ID 1 serves as the baseline model without any semantic guidance. ID 2 integrates only the SFEM module. Specifically, the local semantic features extracted by the LoRA-fine-tuned CLIP image encoder are projected—on a per-unit basis—to match the channel dimension of each SR unit’s output (e.g., 180 channels for RHAG). These projected semantic features are then directly added to the corresponding SR unit outputs. ID 3 builds on ID 2 by incorporating the LMM module. Here, the input channel number of LMM is aligned with the output channel number of the SR unit. ID 4 represents the full SeG-SR model, which includes SFEM, LMM, and the SLM module.

As shown in Table \ref{tab:ablation}, from ID 2 and ID 1, the significant performance improvement upon introducing SFEM indicates that incorporating semantic information substantially enhances SR accuracy. From ID 3 and ID 2, the additional gain after integrating LMM validates the effectiveness of our semantic modulation mechanism. Finally, comparing ID 4 to ID 3, the inclusion of SLM leads to further improvements in both PSNR and SSIM, showing the importance of spatially adaptive semantic localization in guiding the SR process.

\begin{table}[hbt]
\centering
\caption{Ablation Studies on the Key Modules.}
\scalebox{1.10}{
\begin{tabular}{c c c c c c}
\toprule
\multirow{2}{*}{ID}
  & \multicolumn{3}{c}{Setting}
  & \multicolumn{2}{c}{Metric} \\
\cmidrule(lr){2-4} \cmidrule(lr){5-6}
  & SFEM & SLM & LMM
  & SSIM & PSNR (dB) \\
\midrule
1 &      &      &      & 0.7893 & 29.0985 \\
2 & \checkmark &      &      & 0.7947 & 29.2792 \\
3 & \checkmark &      & \checkmark & 0.7950 & 29.2831 \\
4 & \checkmark & \checkmark & \checkmark & \textbf{0.7961} & \textbf{29.3042} \\
\bottomrule
\end{tabular}
}
\label{tab:ablation}
\end{table}

\subsubsection{Parameter analysis of SFEM}
We further conduct ablation experiments on key settings within the SFEM module, focusing on two main factors: the rank number used during LoRA fine-tuning, and the targeted layers within the CLIP image encoder to which LoRA is applied. The results are reported in Table \ref{tab:lora}. In the table, “w/ LoRA” indicates that LoRA fine-tuning is applied. “w/o LoRA” denotes that no LoRA adaptation is used. “QKV” refers to applying LoRA only to the qkv matrices within the attention modules. “QKV+FFN” denotes applying LoRA to both the qkv matrices in the attention blocks and the linear layers in the FFNs.

From the comparison between the first row and the subsequent four rows, we observe a improvement in SR performance when LoRA fine-tuning is used. This result confirms that fine-tuning helps bridge the domain gap between the general-purpose pretraining of CLIP and the specific characteristics of remote sensing scenes.

Among the configurations, the best reconstruction performance is achieved with a rank of 32, as shown in rows 2, 4, and 5. Furthermore, comparing row 3 and row 4, we find that applying LoRA to both the attention and FFN components leads to better SR performance.

\begin{table}[hbt]
\centering
\caption{Ablation of on settings of SFEM.}
\scalebox{1.00}{
\begin{tabular}{l c c c c c}
\toprule
\multirow{2}{*}{LoRA}
  & \multirow{2}{*}{rank}
  & \multirow{2}{*}{QKV}
  & \multirow{2}{*}{QKV+FFN}
  & \multicolumn{2}{c}{Metric} \\
\cmidrule(lr){5-6}
  &  &  &  & SSIM & PSNR (dB) \\
\midrule
w/o.\ LoRA
  & /  & /           & /            & 0.7939 & 29.2521 \\
\midrule
\multirow{4}{*}{w/. LoRA}
  & 16 & \checkmark  & \checkmark            & 0.7942 & 29.2717 \\
  & 32 & \checkmark  &             & 0.7960 & 29.3011 \\
  & 32 & \checkmark  & \checkmark   & \textbf{0.7961} & \textbf{29.3042} \\
  & 64 & \checkmark  & \checkmark   & {0.7958} & {29.2808} \\
\bottomrule
\end{tabular}
}
\label{tab:lora}
\end{table}

\subsubsection{Effectiveness of the detailed designs in SLM}
We further investigate the impact of several key designs in SLM, with the results summarized in Table \ref{tab:slm}. ID 1 denotes only using the per-unit vector \(\mathbf{p}_{i}\) for semantic localization. ID 2 augments ID 1 by adding a randomly initialized global vector \(\mathbf{p}_{g}\) and fuses \(\mathbf{p}_{i}\) with \(\mathbf{p}_{g}\) via gated fusion. ID 3 replaces \(\mathbf{p}_{g}\) directly with CLIP's global semantic feature \(\mathbf{I}_{g}\). ID 4 removes \(\mathbf{p}_{i}\) while retaining \(\mathbf{p}_{g}\) and \(\mathbf{I}_{g}\), fused by MetaNet. ID 5 represents the complete SLM design with all components intact.

From the comparison between ID 1 and ID 2, we observe that introducing the \(\mathbf{p}_{g}\) significantly improves performance, indicating that combining per-unit local representations with a global context enhances semantic localization. Comparing ID 2 and ID 3, we observe an increase in SSIM, suggesting that leveraging \(\mathbf{I}_{g}\) provides more meaningful global information than a randomly initialized \(\mathbf{p}_{g}\), improving the structural fidelity of the reconstruction. The performance difference between ID 3 and ID 5 shows that fusing \(\mathbf{p}_{g}\)  with \(\mathbf{I}_{g}\) results in better semantic localization than using either in isolation. Lastly, the comparison among ID 1, ID 4, and ID 5 demonstrates that the best reconstruction performance is achieved when both local unit-level vectors and scene-aware global semantic features are jointly utilized. This validates the full design of SLM.

\begin{table}[hbt]
\centering
\caption{Ablation Studies on settings of the SLM. }
\scalebox{1.10}{
\begin{tabular}{c c c c c c}
\toprule
\multirow{2}{*}{ID}
  & \multicolumn{3}{c}{Settings}
  & \multicolumn{2}{c}{Metrics} \\
\cmidrule(lr){2-4} \cmidrule(lr){5-6}
  & \(\mathbf{p}_{i}\) & \(\mathbf{p}_{g}\)\,& \(\mathbf{I}_{g}\)\,
  & SSIM & PSNR (dB) \\
\midrule
1 & \checkmark &            &            & 0.7949 & 29.2889 \\
2 & \checkmark & \checkmark &            & 0.7954 & 29.2997 \\
3 & \checkmark &            & \checkmark & 0.7959 & 29.2956 \\
4 &            & \checkmark & \checkmark & 0.7949 & 29.2955 \\
5 & \checkmark & \checkmark & \checkmark & \textbf{0.7961} & \textbf{29.3042} \\
\bottomrule
\end{tabular}
}
\label{tab:slm}
\end{table}

\begin{table}[hbt]
\centering
\caption{Ablation Studies on different architecture of SR Unit and different numbers of SR Unit.}
\scalebox{1.00}{
\begin{tabular}{c l cc cc}
\toprule
\multirow{2}{*}{ID}
  & \multirow{2}{*}{Method}
  & \multicolumn{2}{c}{$\times2$}
  & \multicolumn{2}{c}{$\times4$} \\
\cmidrule(lr){3-4} \cmidrule(lr){5-6}
  & & SSIM & PSNR (dB) & SSIM & PSNR (dB) \\
\midrule
\multirow{4}{*}{1}
  & EDSR  & 0.9383 & 35.5104 & 0.7840 & 28.9274 \\
  & +2 unit    & 0.9389 & 35.6736 & 0.7885 & 29.2717 \\
  & +4 unit    & {0.9397} & \textbf{35.6864} & {0.7892} & \textbf{29.1522} \\
  & +8 unit    & \textbf{0.9399} & 35.6819 & \textbf{0.7895} & 29.1276 \\
\midrule
\multirow{4}{*}{2}
  & RCAN  & 0.9386 & 35.5456 & 0.7860 & 29.0668 \\
  & +2 unit    & 0.9398 & 35.6661 & 0.7897 & 29.1335 \\
  & +5 unit    & {0.9397} & \textbf{35.6819} & \textbf{0.7912} & \textbf{29.1476} \\
  & +10 unit   & \textbf{0.9399} & 35.6724 & 0.7899 & 29.1463 \\
\midrule
\multirow{4}{*}{3}
  & HAT   & 0.9398 & 35.6829 & 0.7893 & 29.0985 \\
  & +2 unit    & 0.9409 & 35.8352 & 0.7934 & 29.2415 \\
  & +3 unit    & 0.9414 & 35.8572 & 0.7944 & 29.2710 \\
  & +6 unit    & \textbf{0.9415} & \textbf{35.8757} & \textbf{0.7961} & \textbf{29.3042} \\
\bottomrule
\end{tabular}
}
\label{tab:baseline}
\end{table}

\begin{figure*}
    \centering    \includegraphics[width=0.95\textwidth]{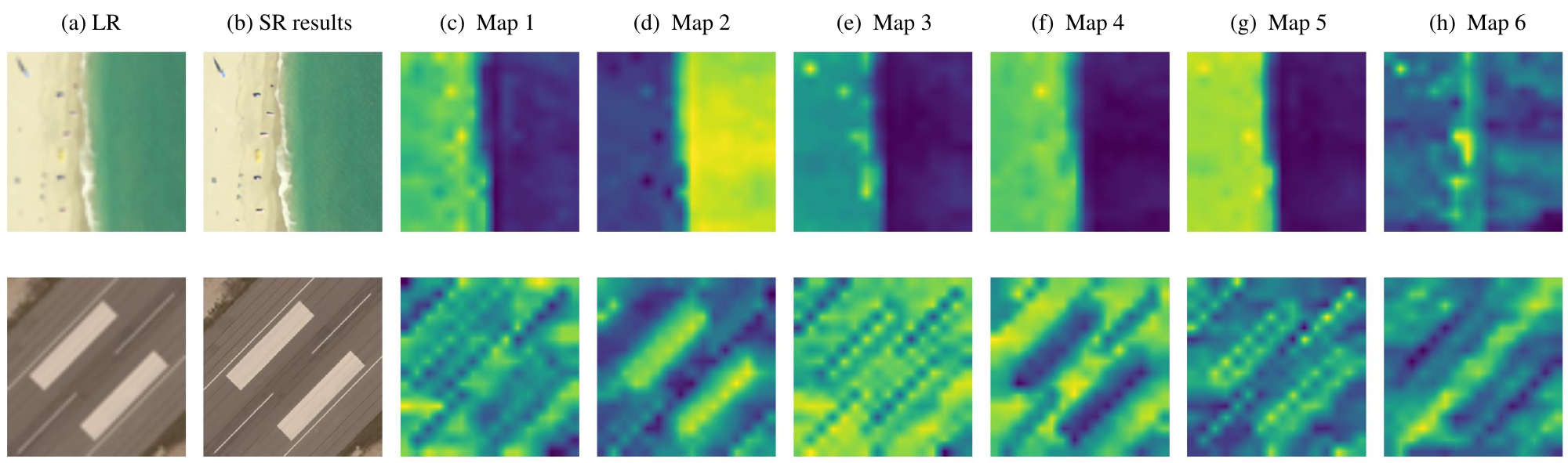}	
    \caption{Visualization of the semantic guidance maps \(\mathbf{M}_i\) used to modulate each SR units during the \(\times\)4 SR task on the UCMerced dataset. The first column shows the input LR images. The second column displays the final SR outputs. The third to eighth columns illustrate the \(\mathbf{M}_i\) corresponding to each of the 6 SR units.}
	\label{fig:uc map}
\end{figure*}

\subsubsection{Function of the semantic guidance map}

We analyze the semantic guidance maps used to modulate each SR unit in SeG-SR, with the visualizations shown in Fig. \ref{fig:uc map}. These results illustrate that the semantic guidance maps effectively localize the regions that each SR unit should focus on during the SR process.

In the first row of Fig. \ref{fig:uc map}, corresponding to a beach scene, the model allocates more semantic guidance to the land areas, as the sea surface contains relatively simpler textures. For example, only the second SR unit is guided to focus on the ocean region, while the first, fourth, and fifth units are directed toward the land. The third unit is guided to attend to small ground objects on the beach, and the sixth unit is focused on the boundary between land and sea.

In the second row of Fig. \ref{fig:uc map}, which depicts a runway scene, the first and third units concentrate on the non-striped ground areas. The fourth unit focuses on the non-white regions, while the second unit attends to two white rectangular shapes in the image. The fifth unit focuses on the line and edges, and the sixth unit focuses on diagonal markings on the runway.

These observations confirm that the semantic guidance maps dynamically distribute semantic focus across SR units, enabling the model to handle diverse scenes more effectively.

\subsubsection{Generalizability of SeG-SR framework}

To validate the generalizability and effectiveness of our SeG-SR framework in enhancing various SR architectures, we apply it to three representative baseline methods: EDSR (CNN-based), RCAN (attention-based), and HAT (Transformer-based). These methods represent distinct architectural paradigms commonly used in SR research. We integrate SeG-SR into each of these methods and conduct experiments on the UCMerced dataset for both \(\times\)2 and \(\times\)4 SR tasks. The results are summarized in Table \ref{tab:baseline}.

Across all three baselines, incorporating the SeG-SR framework consistently leads to performance improvements. This confirms the effectiveness and generalizability of our approach, demonstrating that SeG-SR can work for diverse SR architectures. Moreover, the observed performance gains highlight the critical role of semantic information in advancing the accuracy and fidelity of RSISR. 

For EDSR, the feature extraction network consists of 32 basic residual blocks. Given the simplicity of each individual block, a single block alone has limited capacity of extracting features. Therefore, we group multiple blocks and treat each group as one SR unit. To identify the optimal number of SR units, we divide the 32 blocks into 2, 4, and 8 groups, corresponding to 16, 8, and 4 blocks per unit, respectively. The results are shown in Table \ref{tab:baseline}, ID 1 (Rows 2–4). In both \(\times\)2 and \(\times\)4 SR tasks, 8-unit configuration yields the highest SSIM, while 4-unit configuration achieves the best PSNR. Considering the trade-off between performance and computational efficiency, we recommend the 4-unit configuration as the most practical choice.

For RCAN, the network is composed of 10 Residual Groups (RGs). To evaluate the impact of SR unit number, we experiment with splitting these into 2, 5, and 10 groups. Results are shown in Table \ref{tab:baseline}, ID 2 (Rows 2–4). The 10-unit configuration achieves the highest SSIM in the \(\times\)2 SR task, while the 5-unit configuration consistently outperforms others in PSNR for \(\times\)2 SR and in PSNR and SSIM for \(\times\)4 SR. Therefore, we recommend using 5 SR units for RCAN.

For HAT, the feature extractor contains 6 RHAGs. We test configurations with 2, 3, and 6 SR units. As shown in Table \ref{tab:baseline}, ID 3 (Rows 2–4), the configuration using each RHAG as an individual SR unit (i.e., 6 units) delivers the best performance. This is likely because that each RHAG—composed of stacked Transformer blocks with hybrid attention—has strong feature extraction capabilities, making it well-suited to serve as a single SR unit.

Based on these results, we recommend that each SR unit should possess sufficient feature extraction capacity, and that the number of units should generally be set in the range of 4 to 6 to balance performance and efficiency effectively.

\section{Conclusion}\label{conclusion}

In this paper, we introduce semantic information into the SR network and propose a novel framework for RSISR, termed SeG-SR. The SeG-SR framework consists of three core modules: SFEM, SLM, and LMM.  SFEM leverages a pretrained VLM to extract semantic features from the LR input. SLM generates per-unit semantic localization embeddings, which are used to query the semantic features extracted by SFEM and produce a set of semantic guidance maps specific to each SR unit. LMM uses a set of learnable parameters to modulate the output features of each SR unit based on the corresponding semantic guidance map.

Extensive experiments validate the effectiveness of these three key components. Through comparisons with state-of-the-art methods, we demonstrate the superior performance of SeG-SR in RSISR tasks. Moreover, we evaluate the generalizability of the SeG-SR framework by reconstructing several representative SR architectures using our semantic-guided approach. Across all tested baselines, SeG-SR consistently delivers notable performance gains, confirming its broad applicability. Lastly, we provide a visual analysis of the semantic guidance maps. The results show that the maps successfully guide different SR units to focus on different semantic regions of the image, reinforcing the functional relevance of our design.  However, since SeG-SR incorporates a pretrained vision–language foundation model, it inevitably introduces additional overhead in terms of parameter count and computational cost. Future work may therefore focus on improving its computational efficiency.\color{black}

\ifCLASSOPTIONcaptionsoff
  \newpage
\fi



%
%
%
{\small
  \bibliographystyle{jabbrv_IEEEtran}
  \bibliography{references}
}

\begin{IEEEbiography}[{\includegraphics[width=1in,height=1.25in,clip,keepaspectratio]{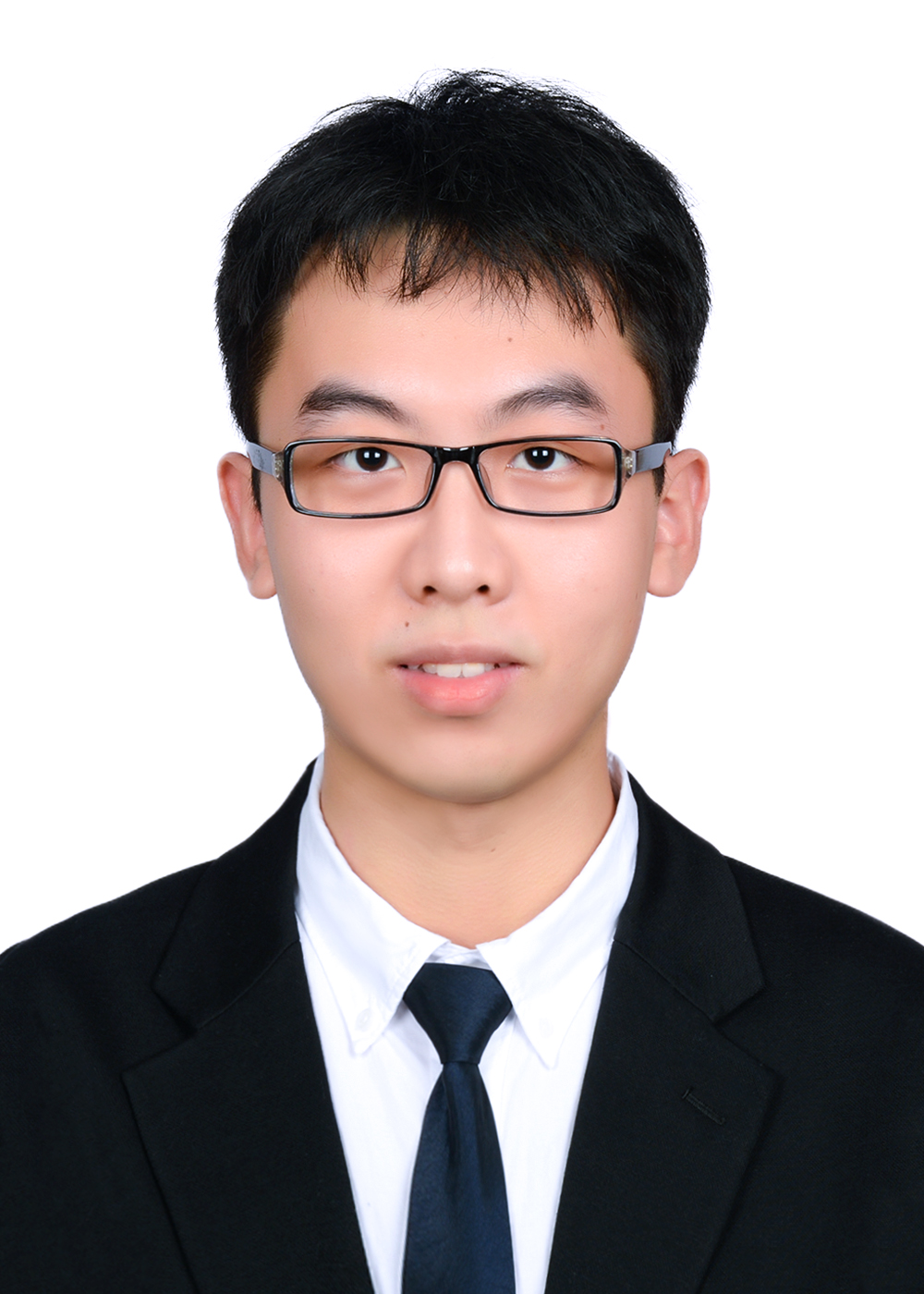}}]
{Bowen Chen}
received his B.S. degree from China University of Petroleum East China, Qingdao, Shandong, China, in 2022. He is currently working toward his doctor's degree in the School of Astronautics, Beihang University. 

His research interests include remote sensing image processing and computer vision.
\end{IEEEbiography}

\begin{IEEEbiography}[{\includegraphics[width=1in,height=1.25in,clip,keepaspectratio]{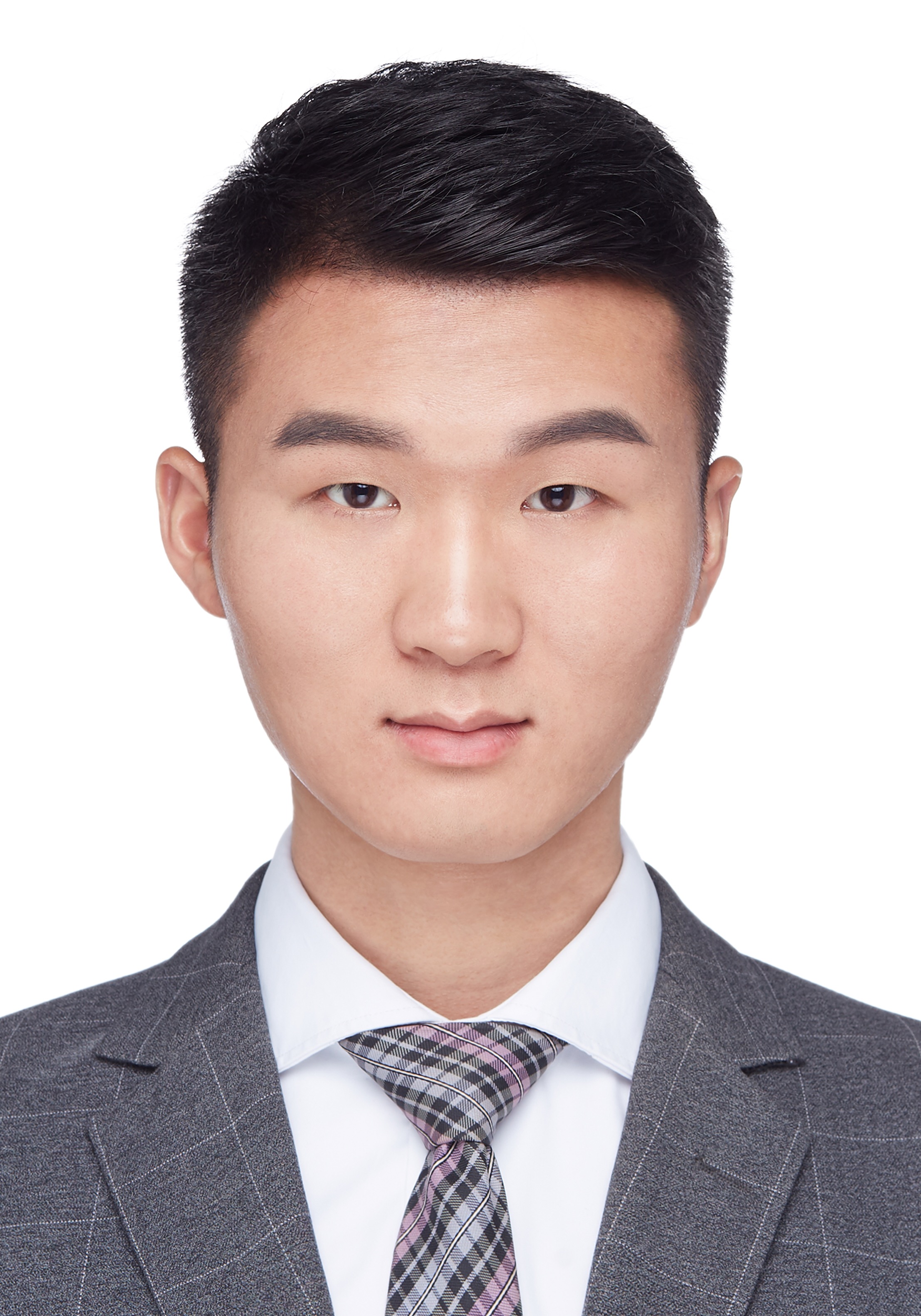}}]{Keyan Chen} 
(Graduate Student Member, IEEE) received his B.S. and M.S. degrees from the School of Astronautics at Beihang University, Beijing, China, in 2019 and 2022, respectively. He is pursuing a Ph.D. in the Image Processing Center, School of Astronautics, Beihang University. His research interests include machine learning, computer vision, and multimodal learning. To date, he has authored over 30 peer-reviewed publications in leading journals and conferences, including \textit{Proceedings of the IEEE}, \textit{IEEE Transactions on Geoscience and Remote Sensing}, and \textit{IEEE/CVF Computer Vision and Pattern Recognition}. His work has been recognized with three ESI Hot Papers and six ESI Highly Cited Papers. His personal website is \url{https://chenkeyan.top}.
\end{IEEEbiography}

\begin{IEEEbiography}[{\includegraphics[width=1in,height=1.25in,clip,keepaspectratio]{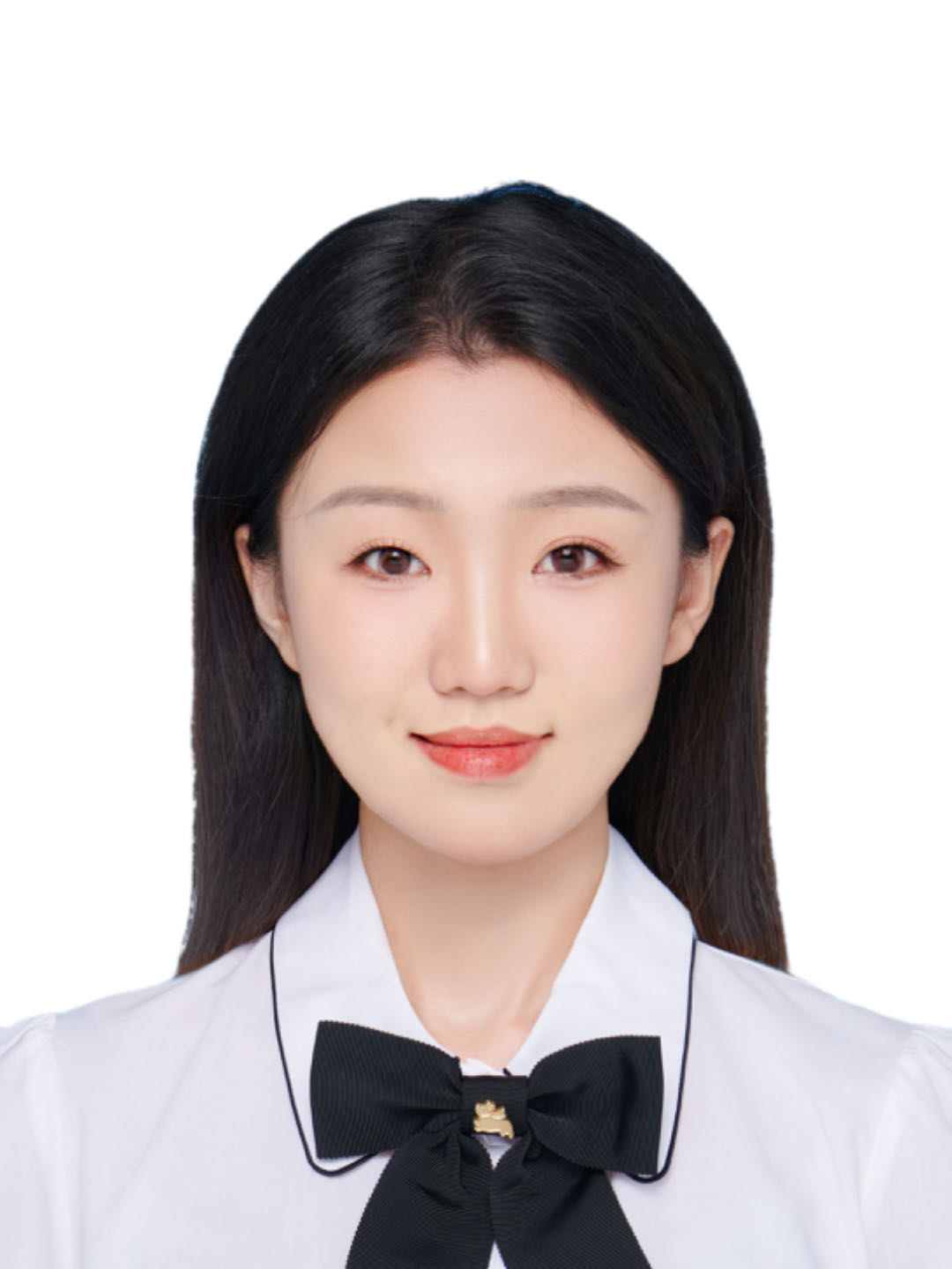}}]
{Mohan Yang} recieved her B.S. degree from Xi'an Jiaotong University , Xi'an, Shanxi,China, in 2024. She is currently working toward her master's degree in the School of Astronautics, Beihang University. Her research interests include satellite video processing and computer vision.
\end{IEEEbiography}

\begin{IEEEbiography}
	[{\includegraphics[width=1in,height=1.25in,clip,keepaspectratio]{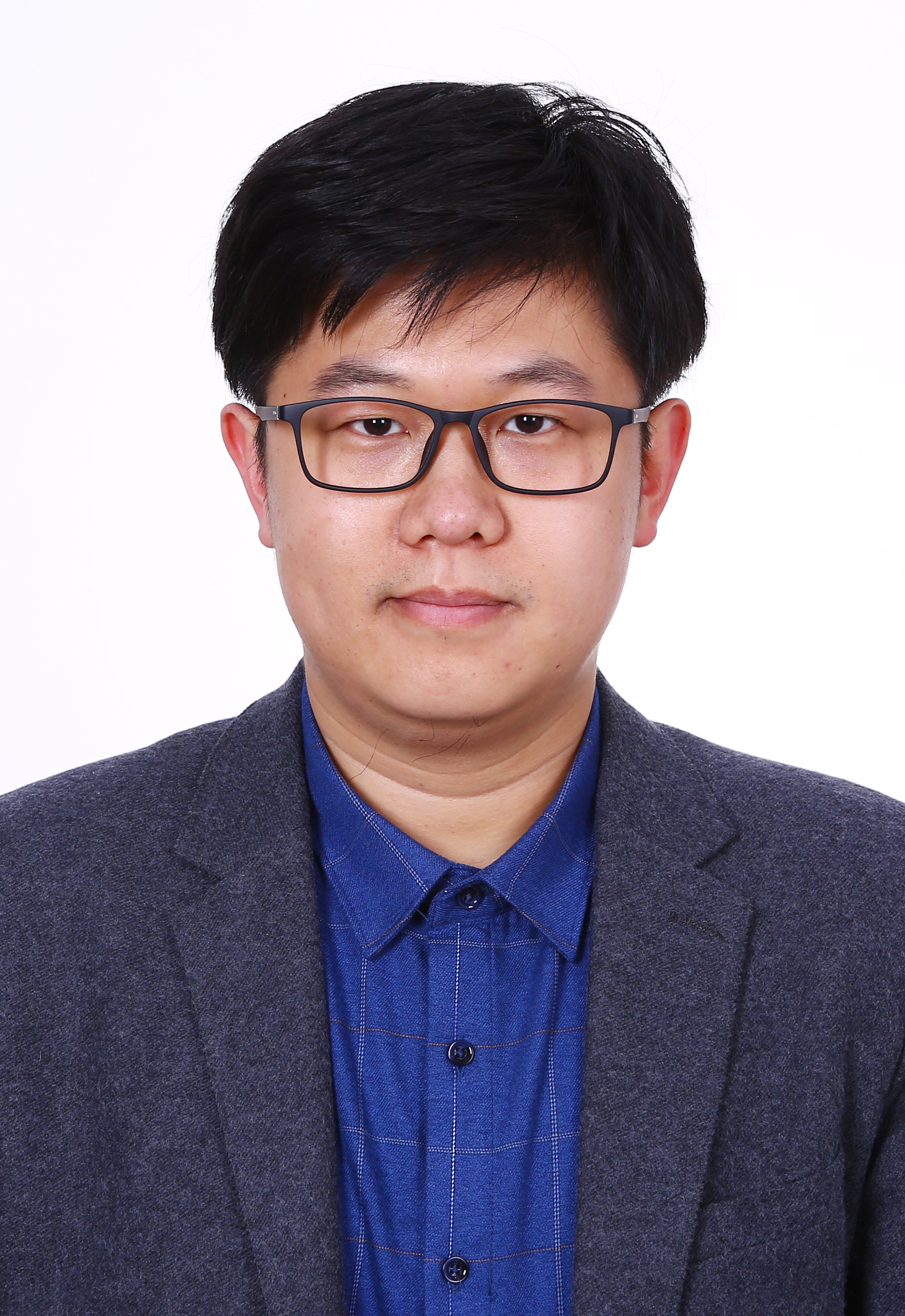}}]
	{Zhengxia Zou} (Senior Member, IEEE) received the B.S. and Ph.D. degrees from Beihang University, Beijing, China, in 2013 and 2018, respectively. He is currently a Professor with the School of Astronautics, Beihang University. From 2018 to 2021, he was a Postdoctoral Research Fellow at the University of Michigan, Ann Arbor, MI, USA. His research interests include remote sensing image processing and computer vision. He has published more than 60 peer-reviewed papers in top-tier journals and conferences, including Nature Communications, Proceedings of the IEEE, IEEE Transactions on Image Processing, IEEE Transactions on Geoscience and Remote Sensing, and IEEE/CVF Computer Vision and Pattern Recognition. Dr. Zou serves as an Associate Editor for IEEE Transactions on Image Processing (TIP). His personal website is \url{https://zhengxiazou.github.io/}.
\end{IEEEbiography}

\begin{IEEEbiography}
[{\includegraphics[width=1in,height=1.25in,clip,keepaspectratio]{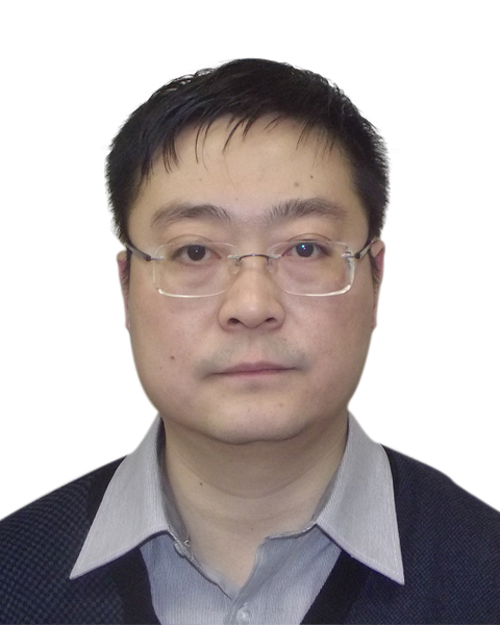}}]
{Zhenwei Shi}(Senior Member, IEEE) is currently a Professor and the Dean of the Department of Aerospace Intelligent Science and Technology, School of Astronautics, Beihang University, Beijing, China. He has authored or coauthored over 300 scientific articles in refereed journals and proceedings. His research interests include remote sensing image processing and analysis, computer vision, pattern recognition, and machine learning. His personal website is \url{https://levir.buaa.edu.cn/}.

Prof. Shi serves as an Editor for IEEE TRANSACTIONS ON GEOSCIENCE AND REMOTE SENSING, Pattern Recognition, ISPRS Journal of Photogrammetry and Remote Sensing, Infrared Physics and Technology, and etc.

\end{IEEEbiography}



\end{document}